\documentclass[runningheads]{llncs}

\usepackage{eccv}

\usepackage{eccvabbrv}

\usepackage{graphicx}
\usepackage{booktabs}

\usepackage[accsupp]{axessibility}  

\usepackage[pagebackref,breaklinks,colorlinks,citecolor=eccvblue]{hyperref}

\usepackage{orcidlink}

\usepackage{url}
\usepackage{booktabs}       
\usepackage{amsfonts}       
\usepackage{nicefrac}       
\usepackage{microtype}      
\usepackage{xcolor}         

\usepackage{xspace}
\usepackage{mathabx}
\usepackage{multirow}
\usepackage{tabularx}
\usepackage{pifont}
\usepackage{graphicx}
\usepackage{arydshln}
\usepackage{blindtext}
\usepackage{bm}
\usepackage{wrapfig}
\usepackage{makecell}

\usepackage{amsmath}

\usepackage{listings}
\usepackage{xcolor}

\definecolor{hollywoodcerise}{rgb}{0.96, 0.0, 0.63}
\definecolor{lasallegreen}{rgb}{0.03, 0.47, 0.19}
\definecolor{hanpurple}{rgb}{0.32, 0.09, 0.98}
\definecolor{green(pigment)}{rgb}{0.0, 0.65, 0.31}
\definecolor{zaffre}{rgb}{0.0, 0.08, 0.66}

\definecolor{codegreen}{rgb}{0,0.6,0}
\definecolor{codegray}{rgb}{0.5,0.5,0.5}
\definecolor{codepurple}{rgb}{0.58,0,0.82}
\definecolor{backcolour}{rgb}{0.95,0.95,0.92}

\lstdefinestyle{mystyle}{
    backgroundcolor=\color{backcolour},
    commentstyle=\color{codegreen},
    keywordstyle=\color{magenta},
    numberstyle=\tiny\color{codegray},
    stringstyle=\color{codepurple},
    basicstyle=\ttfamily\footnotesize,
    breakatwhitespace=false,
    breaklines=true,
    captionpos=b,
    keepspaces=true,
    numbers=right,
    numbersep=5pt,
    showspaces=false,
    showstringspaces=false,
    showtabs=false,
    tabsize=2
}
\newcommand{\citep}[1]{\cite{#1}}
\newcommand{\ub}[1]{\underline{\textbf{{#1}}}}

\newcommand{\mt}[1]{{\color{black}{{{#1}}}}}
\newcommand{\rt}[1]{{\color{black}{{{#1}}}}}

\begin{document}

\title{UniINR: Event-guided Unified Rolling Shutter Correction, Deblurring, and Interpolation}

\titlerunning{Abbreviated paper title}

{
\author{Yunfan LU\inst{1}\orcidlink{0000-0002-7371-3189} \and
Guoqiang Liang\inst{1}\orcidlink{0000-0003-4790-7075} \and
Yusheng Wang\inst{2}\orcidlink{0000-0002-8665-3437} \and  \\
Lin Wang\inst{1} \and
Hui Xiong$^*$\inst{1}\orcidlink{0000-0001-6016-6465}}

\authorrunning{Y.F. Lu et al.}
\institute{ AI Thrust, HKUST(GZ);  University of Tokyo \\
\email{\{ylu066,gliang041\}@connect.hkust-gz.edu.cn},\\
\email{wang@race.t.u-tokyo.ac.jp}, \email{\{linwang,xionghui\}@ust.hk}}
}

\begingroup
\renewcommand\thefootnote{*}
\footnotetext{Corresponding author.}
\endgroup

\maketitle
\begin{abstract}
Video frames captured by rolling shutter (RS) cameras during fast camera movement frequently exhibit RS distortion and blur simultaneously.
Naturally, recovering high-frame-rate global shutter (GS) sharp frames from an RS blur frame must simultaneously consider RS correction, deblur, and frame interpolation.
A naive way is to decompose the whole process into separate tasks and cascade existing methods; however, this results in cumulative errors and noticeable artifacts.
Event cameras enjoy many advantages, \eg, high temporal resolution, making them potential for our problem. To this end, we propose the \textbf{first} and novel approach, named \textbf{UniINR}, to recover arbitrary frame-rate sharp GS frames from an RS blur frame and paired events.
Our key idea is unifying spatial-temporal implicit neural representation (INR) to directly map the position and time coordinates to color values to address the interlocking degradations.
Specifically, we introduce spatial-temporal implicit encoding (STE) to convert an RS blur image and events into a spatial-temporal representation (STR).
To query a specific sharp frame (GS or RS), we embed the exposure time into STR and decode the embedded features pixel-by-pixel to recover a sharp frame.
Our method features a lightweight model with only \textbf{0.38$M$} parameters, and it also enjoys high inference efficiency, achieving \textbf{2.83}$ms/frame$ in $31 \times$ frame interpolation of an RS blur frame.
Extensive experiments show that our method significantly outperforms prior methods.
Code is available at \url{https://github.com/yunfanLu/UniINR}.
\end{abstract}

\section{Introduction}
Most consumer-level cameras based on CMOS sensors rely on a rolling shutter (RS) mechanism. These cameras dominate the market owing to their benefits, \eg, low power consumption, simple design, and high sensitivity~\citep{janesick2009fundamental,lu2022all}.
In contrast to the global shutter (GS) cameras, RS cameras capture pixels row by row; thus, the captured frames often suffer from obvious spatial distortions (\eg, stretch) and blur under fast camera/scene motion.
Theoretically, an RS frame can be formulated as a row-wise combination of sequential GS frames within the exposure time~\citep{fan2021inverting,fan2023rolling}.
Naively neglecting the RS effect often hampers the performance in many real-world applications ~\citep{hedborg2012rolling,lao2020rolling,zhong2021towards,zhou2022evunroll}.
In this regard, it is meaningful to \textit{recover high-frame-rate sharp GS frames from a single RS blur frame} as the restored GS frames can directly facilitate many downstream tasks.
Intuitively, achieving this goal often requires considering \textbf{RS correction}, \textbf{deblurring}, and \textbf{frame interpolation} simultaneously.
However, tackling this task is nontrivial because multiple degradations, such as RS distortion, motion blur, and temporal discontinuity~\citep{meilland2013unified,su2015rolling,wang2022efficient}, often co-exist for CMOS cameras~\citep{zhong2021towards}.
The co-existence of various image degradations complicates the whole GS frame restoration process.
A naive way is to decompose the whole process as separate tasks and cascade existing image enhancement networks, which can result in cumulative errors and noticeable artifacts.
For example, a simple consideration of cascading a frame interpolation network~\citep{bao2019depth} with an RS correction network produces degraded results, as previously verified in~\citep{naor2022combining}.

\begin{figure*}[t]
\centering
\includegraphics[width=1\linewidth]{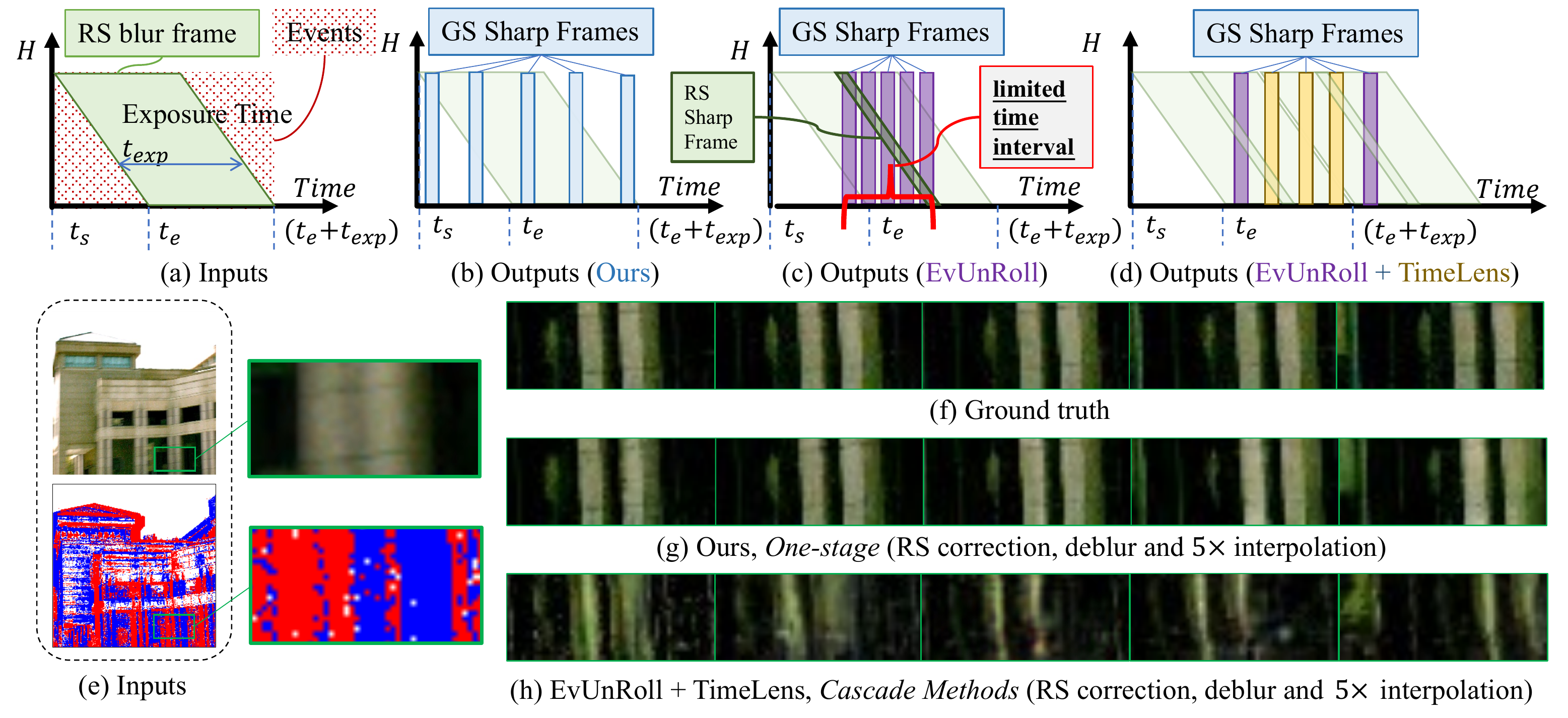}
\caption{Inputs and the outputs of our method, EvUnRoll~\cite{zhou2022evunroll}, and EvUnRoll~\cite{zhou2022evunroll}+TimeLens~\cite{tulyakov2021time}. Inputs are shown in (a), which includes an RS blur frame and events. $t_s$ and $t_e$ are the start and end timestamps of RS, and $t_{exp}$ is the exposure time.
Our outputs are shown in (b), which is a sequence of GS sharp frames during the \textbf{\underline{whole exposure time($ts$, $t_e+t_{exp}$)}} of the RS blur image.
(c) shows outputs of EvUnRoll, which can only recover the GS sharp frames
in a \textbf{\underline{limited time interval}} (\textcolor{red}{red interval}) instead of the whole exposure time of the RS blur frame.
(d) shows outputs of cascade methods of EvUnRoll+TimeLens. \textit{More details are in Supp. Mat.} }
\label{fig:inputs_and_outputs}
\centering
\end{figure*}

Event cameras offer several advantages, \eg, high-temporal resolution, which make them suitable for various image restoration tasks~\citep{wang2020event,liang2024towards,tulyakov2021time,lu2023self,lu2024event,zheng2023deep}.
eSL-Net~\citep{wang2020event} proposes an event-guided sparse learning framework to simultaneously achieve image super-resolution, denoising, and deblurring.
TimeLens~\citep{tulyakov2021time} integrates two branches to boost the performance of the video frame interpolation (VFI).
DeblurSR~\citep{song2023deblursr} and E-CIR~\citep{song2022cir} take advantage of the high temporal resolution of events by converting a blurry frame into an explicit time-to-intensity function to handle VFI and deblurring.
EvUnRoll~\citep{zhou2022evunroll}, SelfUnroll~\citep{wang2023self} and EvShutter~\citep{erbach2023evshutter} leverage events to enhance RS correction by accounting for nonlinear motion.
{\textbf{However, these methods cannot recover arbitrary frame-rate sharp GS frames from a single RS blur frame.}}
Fig.~\ref{fig:inputs_and_outputs} (c) illustrates that EvUnRoll~\cite{zhou2022evunroll}, limited by its pre-trained deblurring network, only recovers sharp GS frames in a \ub{limited time interval} (\textcolor{red}{red interval}).
Combining event-guided RS correction (EvUnroll~\cite{zhou2022evunroll}) with a VFI model (TimeLens~\cite{tulyakov2021time}) for high-frame-rate GS frame recovery leads to significant artifacts, as shown in Fig.~\ref{fig:inputs_and_outputs}~(h).

In this paper, we make the \textbf{first} attempt to propose a novel yet efficient framework, dubbed \textbf{UniINR}, that can {\textbf{recover arbitrary frame-rate sharp GS frames from an RS blur frame and events}}.
Our key idea is to employ sparse learning to generate a spatial-temporal \textit{implicit neural representation (INR) that can directly map the position and time coordinates to color values to address the co-existence of degradations in CMOS cameras}.
Thus, UniINR seamlessly integrates RS correction, deblurring, and interpolation in one unified process, exemplifying our approach's cohesive capability.
Specifically, we formulate the task~\textemdash~ recovering high-frame-rate sharp GS frames from an RS blur frame and paired events~\textemdash~ as a novel \textit{estimation} problem, defined as a function, $F(\bm{x},t, \theta)$. Here, $\bm{x}$ denotes the pixel position $(x,y)$ of an image, $t$ denotes the timestamp during the exposure time, and $\theta$ denotes the function's parameters.

Our framework consists of three parts:
\rt{
\textbf{(1)} spatial-temporal implicit encoding (STE),
\textbf{(2)} exposure time embedding (ETE), and
\textbf{(3)} pixel-by-pixel decoding (PPD).
Specifically, STE first extracts a spatial-temporal representation (STR) as $\theta$ from events and an RS blur frame.
To query a specific sharp frame of RS or GS pattern, we then model the exposure information as a temporal tensor $T$ in ETE.
Finally, PPD leverages an MLP to decode sharp frames from the STR and the temporal tensor $T$, allowing for the generation of a sharp frame at any given exposure pattern (\eg, RS or GS).
}
One notable advantage of our approach is its \ub{high efficiency}, as it only requires using the STE once, regardless of the number of interpolation frames.
Hence, while the frame interpolation multiples increase linearly from $1\times$ to $31 \times$, the practical time required rises from $31 ms$ to $86 ms$, showcasing the non-linear growth in time consumption.
Specifically, at $31 \times$ interpolation, each frame's processing time is merely \textbf{2.8}$ms$, whereas the cascading approach (EvUnRoll~\cite{zhou2022evunroll} + TimeLens~\cite{tulyakov2021time}) requires more than $177 ms$.
Another advantage of UniINR is its lightweight model, boasting only \textbf{0.379M} parameters, a benefit from our unified approach.
We conduct a thorough evaluation of our proposed method, including both quantitative and qualitative analyses.
Results from a variety of RS and blur settings, including both simulated and real-world datasets, demonstrate that our approach outperforms existing methods in RS correction, deblur, and VFI.

In general, we present three key contributions:
\textbf{(1) Novel question}: We pioneer the simultaneous exploration of RS correction, deblurring, and VFI, charting novel territory in event-guided imaging.
\textbf{(2) Innovative framework}: We propose a one-stage framework employing a unified INR for multi-task processing.
\textbf{(3) Efficient performance}: Our approach performance significantly surpasses prior methods with performance and efficiency in model size and speed.

\section{Related Works}
\noindent \textbf{Event-guided RS correction, deblurring, and interpolation:}
Given the high temporal resolution of events,
much prior research has employed events into the task of RS correction~\cite{zhou2022evunroll,erbach2023evshutter,wang2023self,lu2023self}, deblurring~\citep{sun2022event,wang2020event,shang2021bringing,kim2022event},  and VFI~\cite{tulyakov2021time,xu2021motion,song2022cir,song2023deblursr,haoyu2020learning,pan2019bringing,zhang2022unifying,lin2020learning}.
Despite this, existing research often tackles one or two of these tasks in isolation, overlooking the interconnected impacts among them.
To structure our discussion, we categorize the most pertinent research into two main groups.
\noindent\textbf{RS correction + Deblurring:}
Wang \etal~\cite{wang2023self} and Lu \etal~\cite{lu2023self} have explored the use of events to guide the correction of RS frames.
However, their approaches fall short of addressing simultaneous RS and blur, thereby limiting their applicability.
EvUnroll~\citep{zhou2022evunroll} and EvShutter~\cite{erbach2023evshutter} employ events for RS correction and take into account the influence of blur.
\textit{However, their approach of using a two-stage approach for initial deblurring followed by RS correction increases the model size and the cumulative errors.}
\noindent\textbf{Video Frame Interpolation:}
Event-guided VFI stands out as a pivotal area of research, yet most recent works only focus on input frames with GS patterns.
This research branches into two streams: sharp frame-based VFI~\cite{tulyakov2021time,lu2024hr,tulyakov2022time} and VFI attentive to blur effects~\cite{xu2021motion,song2022cir,song2023deblursr,haoyu2020learning}.
TimeLens~\cite{tulyakov2021time} exemplifies the former by utilizing events to predict nonlinear motion, thereby outperforming frame-based methods~\cite{jiang2018super,xu2019quadratic,bao2019depth,kalluri2023flavr}.
On the other hand, works like E-CIR~\citep{song2022cir} and DeblurSR~\citep{song2023deblursr} exemplify the latter by crafting an explicit time-to-intensity mapping, enabling frame interpolation even among blurring.
\textit{However, these methods overlook the impact of RS distortion, leading to performance degradation when dealing with spatially distorted and RS blur frames.}
Recently, NIRE \citep{zhang2023neural} addressed these three tasks.
However, NIRE~\citep{zhang2023neural} mainly focused on a single model's performance on each task separately, lacking a thorough exploration of all three simultaneously—a key difference from our motivation and framework.

\noindent \textbf{Frame-based RS correction and deblurring:}
\noindent\textbf{RS Correction:}
Frame-based methods~\cite{liu2020deep,zhong2021towards,fan2021sunet,zhong2020efficient,zhong2021towards,fan2021inverting,fan2022context} lack motion information between frames, making it hard to model nonlinear motion.
RSSR~\citep{fan2021inverting,fan2023rolling} and CVR~\citep{fan2022context} represent these methods, with RSSR introducing bi-directional undistortion flows, and CVR enhancing motion and aggregating context to estimate and refine GS frames.
\textit{However, RSSR~\citep{fan2021inverting,fan2023rolling} and CVR~\citep{fan2022context} are constrained by their assumption of linear motion for inter-frame dynamics, limiting their capacity to tackle complex motion.}
\noindent\textbf{RS Correction + Deblurring:}
JCD~\citep{zhong2021towards} proposes the first pipeline that employs warping and deblurring branches to address the RS distortion and motion blur.
\textit{However, JCD also built upon the assumption of linear motion derived from DeepUnrollNet~\citep{liu2020deep}, encounters a significant performance degradation in complex scenarios involving non-linear motion~\citep{zhou2022evunroll}.}

\noindent \textbf{Implicit Neural Representation (INR):}
INR~\citep{wang2021nerf,sitzmann2020siren,chen2021liif,chen2022videoinr,lu2023learning} is proposed for parameterized signals (images, video or audio) in the coordinate-based representation, inspiring some researchers to explore the potential of INR in low-level vision tasks.
LIIF~\citep{chen2021liif} represents images as high-dimensional tensors, enabling super-resolution (SR) at any scale.
This approach is followed by VideoINR~\citep{chen2022videoinr} and MoTIF~\cite{chen2023motif}, which extend INR to videos, thereby unifying SR and VFI.
EG-VSR~\citep{lu2023learning} incorporates events into the learning of INR to achieve random-scale video SR.
\rt{
Differently, our approach utilizes an INR that directly maps position and time coordinates to color/intensity values, effectively tackling the \textbf{multifaceted degradations}.
This enables our method to \textbf{simultaneously address} RS correction, deblurring, and interpolation, offering a unified solution to these intertwined challenges.
}

\section{Methodology}

\begin{figure*}[t]
\centering
\includegraphics[width=\linewidth]{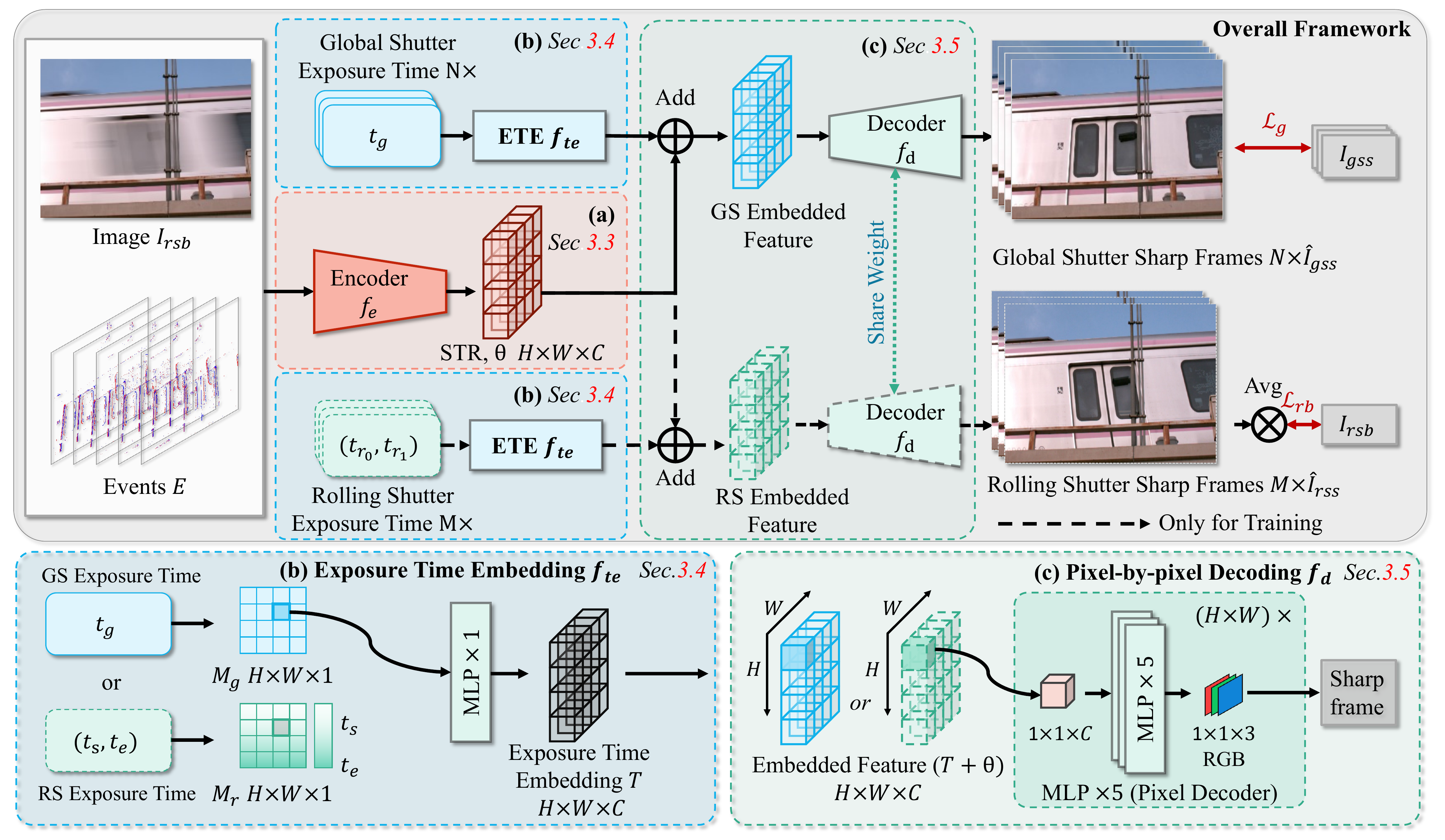}
\caption{An overview of our framework, which consists of three parts, \textbf{(a)} the Spatial-Temporal Implicit Encoding (STE), \textbf{(b)} Exposure Time Embedding (ETE), and \textbf{(c)} Pixel-by-pixel decoding (PPD). Details of STE, ETE, and PPD are described in Sec.~\ref{sec:STE}, Sec.~\ref{sec:ETE}, and Sec.~\ref{sec:PPD}. The inputs are an RS blur frame $I_{rsb}$ and events, and the outputs are a sequence of GS frames and RS frames. RS frames are predicted only in training.}
\label{fig:over_all_framework}
\centering
\end{figure*}

\subsection{Problem Definition and Analysis}
\label{sec:Problem definition}
We formulate the task~\textemdash~\textbf{recovering arbitrary frame-rate sharp GS frames from an RS blur frame and paired events}~\textemdash~ as a novel estimation problem, defined as a function, $F(\bm{x},t,\theta)$. Here, $\bm{x}$ denotes the pixel position $(x,y)$ of a frame with a resolution of $H\times W$, $t$ denotes the timestamp during the exposure time, and $\theta$ denotes the parameters.
The intuition behind this formulation is that there exists a relationship between the RS blur/sharp frame and the GS blur/sharp frame. We now describe it.

By defining a function $F(\bm{x},t,\theta)$ mapping the pixel position $\bm{x} = (x,y)$ and timestamp $t$ to intensity or RGB value, we can obtain a GS sharp frame by inputting the desired timestamp $\hat{t}$ during the exposure time to the function, which can be formulated as:
\begin{equation}
I_{g,\hat{t}}= F(\bm{x},\hat{t}, \theta)
\label{eq:latent-function}
\end{equation}
As an RS frame can be formulated as a row-wise combination of sequential GS frames within the exposure time~\citep{fan2021inverting,fan2023rolling}, we can assemble an RS sharp frame $I_{r,t_s,t_e}$ from a sequence of GS sharp frames row by row given the RS start time $t_s$ and the end time $t_e$, as shown in Fig.~\ref{fig:inputs_and_outputs}.
That is, the $h$-th row of an RS frame is the same as the $h$-th row of a GS frame at $t^h_s$, and the exposure start timestamp of the $h$-th row of an RS frame is $t_s^h=t_s + h\times (t_e-t_s) / H$.
Therefore, we can formally describe an RS sharp frame as Eq.~\ref{eq:rolling_shutter_t0_to_t1_all}, where $[h]$ refers to $h$-th row.
\begin{equation}
    I_{r,t_s,t_e} = \left\{ F\left(\bm{x},t_s^h, \theta \right)[h] , h\in [0, H) \right\}
    \label{eq:rolling_shutter_t0_to_t1_all}.
\end{equation}
\rt{
In principle, a blur frame can be regarded as the temporal average of a sequence of sharp frames~\citep{Nah_2017_CVPR,zhang2020deblurring}.
Thus, a GS blur frame $I_{g,t_g,t_{exp}}$, where $t_g$ is the exposure start timestamp and $t_{exp}$ is the exposure time, can be expressed as the average of a sequence of GS sharp frames during the exposure time $t_{exp}$.
This concept can be formulated as Eq.~\ref{eq:global_shutter_blur_frame_int_sum},
where $N$ is the length of the GS frame sequence and the CRF~\cite{grossberg2003space} is omitted for simplicity.
}
\begin{equation}
I_{g, t, t_{exp}}=\frac{1}{t_{exp}} \int_{t}^{t+t_{exp}} F(\bm{x},t, \theta)dt 	\approx \frac{1}{N} \sum_{i=0}^N I_{g, t_0+i\times t_{exp} / N},
\label{eq:global_shutter_blur_frame_int_sum}
\end{equation}
With the above formulation, an RS blur frame $I_{r, t_s \to t_e, t_{exp}}$ can thus be described based on the RS start time $t_s$, RS end time $t_e$, and exposure time of each scan line $t_{exp}$, as depicted in Fig.~\ref{fig:inputs_and_outputs} (a).
According to Eq.~\ref{eq:rolling_shutter_t0_to_t1_all} and Eq.~\ref{eq:global_shutter_blur_frame_int_sum}, the $h$-th row of an RS blur frame can be described as the temporal average of the $h$-th row in a sequence of GS sharp frames, which can be written as follows:
\begin{equation}
\begin{array}{rl}
     I_{r, t_s \to t_e, t_{exp}} & =\left\{\frac{1}{t_{exp}} \int_{t_s^h}^{t_s^h+t_{exp }} F\left(\bm{x}, t, \theta \right)[h] d t, h \in[0, H)\right\} \\
                                 & \approx \left\{\frac{1}{N} \sum_{i=0}^N I_{g, t_s+i\times t_{exp} / N}[h], h \in[0, H)\right\}.
\end{array}
\label{eq:rolling_shutter_blur}
\end{equation}

An event stream $E$ consists of a set of event $e=(x,y,t,p)$, where each event is triggered and recorded with the polarity $p$ when the logarithmic brightness change at pixel $(x,y)$ exceeds a certain threshold $C$, which can be approximated as the differential of $F(\bm{x},t,\theta)$ with respect to the time dimension. \textit{For details about the principle of event cameras, refer to the Suppl. Mat.}

To use events $E$ as guidance,  we need to address three challenges to estimate the mapping function $F(\bm{x},t,\theta)$:
{
{\color{blue}{(I})} how to find a function $f_e$ to encode the input RS blur image and events to $\theta$ of the mapping function $F(\bm{x},t,\theta)$;
{\color{blue}{(II})} how to find a function $f_{te}$ to represent the exposure information of desired RS or GS sharp frames as $t$ of the mapping function $F(\bm{x},t,\theta)$;
{\color{blue}{(III})}) how to find a function $f_d$ to eliminate the need to input position information of desired RS or GS sharp frames as $p$ of the mapping function $F(\bm{x},t,\theta)$. Therefore, our goal is to estimate $f_e$, $f_{te}$, and $f_d$ in order to get a mapped result, which can be formulated as:
}

\begin{equation}
\begin{tikzpicture}[baseline=(current bounding box.center)]
    \node[inner sep=0] (formula) at (0,0) {$I\!\!=\!\!F(\bm{x}, t, \theta)
\!\!=\!\!F(\bm{x}, t, f_e(E, I_{rsb}))
\!\!=\!\!F(\bm{x}, f_{te}(t), f_e(E, I_{rsb}))
\!\!=\!\!f_d(f_{te}(t), f_e(E, I_{rsb})).$};
    \begin{scope}[overlay]
        \draw[-,blue] (formula.base west) ++(1.3em,0.9em) -- ++(+1.2,0);
        \draw[->,blue] (formula.base west) ++(2.6em,0.9em) -- ++(0,0.2) -- ++(+1.8,0) node[midway,fill=white] {(I)} -- ++(0,-0.2);
        \draw[-,blue] (formula.base west) ++(5.9em,0.9em) -- ++(+2.5,0);
        \draw[->,blue] (formula.base west) ++(11.3em,0.9em) -- ++(0,0.2) -- ++(+1.8,0) node[midway,fill=white] {(II)} -- ++(0,-0.2);
        \draw[-,blue] (formula.base west) ++(14.4em,0.9em) -- ++(+3.2,0);
        \draw[->,blue] (formula.base west) ++(22.3em,0.9em) -- ++(0,0.2) -- ++(+1.8,0) node[midway,fill=white] {(III)} -- ++(0,-0.2);
        \draw[-,blue] (formula.base west) ++(25.2em,0.9em) -- ++(+2.9,0);
    \end{scope}
\end{tikzpicture}
\label{eq:total_equation}
\end{equation}

In the following section, we describe our framework based on Eq.~\ref{eq:total_equation} by substantiating $f_e$, $f_{te}$, and $f_d$.

\subsection{Overview Proposed Framework}
An overview of our UniINR framework is depicted in Fig.~\ref{fig:over_all_framework}, which takes an RS blur image $I_{rsb}$ and paired events $E$ as inputs and outputs $N$ sharp GS frames $\{I_{gss}\}_{i=0}^N$ with a high-frame-rate.
To substantiate the defined functions $f_e$, $f_{te}$, and $f_d$, as mentioned in Sec.~\ref{sec:Problem definition}, our proposed framework consists of three components:
\textbf{(1)} Spatial-Temporal Implicit Encoding (STE),
\textbf{(2)} Exposure Time Embedding (ETE), and
\textbf{(3)} Pixel-by-pixel Decoding (PPD).
Specifically, we first introduce an STE to encode the RS blur frame and events into a spatial-temporal representation (STR) (Sec.~\ref{sec:STE}).
To provide exposure temporal information for STR, we embed the exposure start timestamp of each pixel from the GS or RS by ETE. (Sec.~\ref{sec:ETE}).
Lastly, the PDD module adds ETE to STR to generate RS or GS sharp frames (Sec.~\ref{sec:PPD}). We now describe these components in detail.

\subsection{Spatial-Temporal Implicit Encoding (STE)}
\label{sec:STE}
\rt{
Based on the analysis in Sec.~\ref{sec:Problem definition}, we conclude that the RS blur frame $I_{rsb}$ and events $E$ collectively encompass the comprehensive spatial-temporal information during the exposure process.
In this section, we aim to extract a spatial-temporal implicit representation $\theta$ that can effectively capture the spatial-temporal information from the RS blur frame $I_{rsb}$ and events $E$.

To achieve this, we need to consider two key factors: (1) extracting features for the multi-task purpose and (2) estimating motion information.
For the first factor, we draw inspiration from eSL-Net~\citep{wang2020event}, which effectively utilizes events to simultaneously handle deblur, denoise, and super-resolution tasks. Accordingly, we design a sparse-learning-based backbone for the encoder.
Regarding the second factor, previous works~\citep{fan2021inverting,fan2022context,fan2023rolling} use optical flow for motion estimation in RS correction, deblurring, and VFI.
However, optical flow estimation is computationally demanding~\citep{gehrig2021raft,zhu2019unsupervised,sun2018pwc}, making it challenging to incorporate it into the multiple task framework for RS cameras due to the complex degradation process.

As an efficient alternative, we propose to estimate a video STR by the sparse-learning-based backbone. And the STR stores motion information and directly maps time and coordinates to RGB values, as shown in Eq.~\ref{eq:latent-function}.
In practice, we adopt a 3D tensor with a shape of $H\times W\times C$ as the STR $\theta$, which can effectively address the interlocking degradations encountered in the image restoration process with a sparse-learning-based backbone, as formulated as $\theta = f_e(E, I_{rsb})$ in Eq.~\ref{eq:total_equation}.
\textit{More details in the Suppl. Mat.}
}

\subsection{Exposure Time Embedding (ETE)\label{sec:ETE}}
As depicted in Fig.~\ref{fig:over_all_framework} (b), the objective of the ETE module is to incorporate the exposure time of either a rolling shutter (RS) frame ($t_s, t_e$) or a global shutter (GS) frame ($t_g$) by employing an MLP layer, resulting in the generation of a temporal tensor $T$.
To achieve this, we design an ETE module, denoted as $f_{te}$, which takes the GS exposure time $t_g$ as input and produces the GS temporal tensor $T_g = f_{te}(t_g)$.
Similarly, for RS frames, $T_r = f_{te}(t_{r_s}, t_{r_e})$ represents the RS temporal tensor, which is only used in training.
The process begins by converting the exposure process information into a timestamp map, with a shape of $H\times W\times 1$. Subsequently, the timestamp map is embedded by increasing its dimensionality to match the shape of the STR. This embedding procedure allows for the integration of the exposure time information into the STR representation.
We now explain the construction of timestamp maps for both GS and RS frames and describe the embedding method employed in our approach.
\\
\noindent\textbf{GS Timestamp Map:}
In GS sharp frames, all pixels are exposed simultaneously, resulting in the same exposure timestamps for pixels in different positions.
Given a GS exposure timestamp $t_g$, the GS timestamp map $M_{g}$ can be represented as $M_{g}[h][w] = t_{g}$, where $h$ and $w$ denote the row and column indices, respectively.
\\
\noindent\textbf{RS Timestamp Map:}
According to in Sec.~\ref{sec:Problem definition}, pixels in RS frames are exposed line by line, and pixels in different rows have different exposure start timestamps.
Given RS exposure information with start time $t_s$ and RS end time $t_e$,
the RS timestamp map can be represented as $M_{r}[h][w] = t_s + (t_e - t_s)\times h / H$, where $h$, $w$, $H$ denote the row and column indices and height of the image, respectively.
\\
\noindent\textbf{Time Embedding:} The timestamp maps, $M_r$ and $M_g$, represent the timestamps of each pixel in a specific frame (RS or GS) with a shape of $H \times W \times 1$.
\mt{
However, the timestamp map is a high-frequency variable and can pose challenges for learning neural networks~\citep{vaswani2017attention}.
Some approaches~\citep{vaswani2017attention,wang2021nerf} propose a combination function of sine and cosine to encode the positional embedding.
}
In this paper, we utilize a one-layer MLP to increase the dimension for embedding.
The whole embedding process is formulated as $T_g = f_{te}(t_g)$ for GS frames, and $T_r = f_{te}(t_{r_s}, t_{r_e})$ for RS frames, as depicted in Fig.~\ref{fig:over_all_framework} (b).
The MLP consists of a single layer that maps the timestamp map $M_r$ or $M_g$ to the same dimension $H\times W\times C$ as the spatial-temporal representation (STR) $\theta$, as described in Sec.~\ref{sec:STE}.

\subsection{Pixel-by-pixel Decoding (PPD)}
\label{sec:PPD}
\mt{
As shown in Fig.~\ref{fig:over_all_framework} (c), the goal of PPD is to efficiently query a sharp frame from STR $\theta$ by the temporal tensor $T$.
It is important that the encoder is invoked only once for $N$ times interpolation, while the decoder is called $N$ times.
Therefore, the efficiency of this query is crucial for the overall performance.
The query's inputs $\theta$ capture the global spatial-temporal information, and $T$ captures the temporal information of the sharp frame (GS or RS).
Inspired by previous works~\citep{mildenhall2021nerf,chen2021liif}, we directly incorporate the temporal tensor $T$ into the STR $\theta$ to obtain an embedded feature with a shape of $H\times W\times C$ for each query.
This additional embedded feature combines the global spatial-temporal information with the local exposure information, enabling straightforward decoding to obtain a sharp frame.
To avoid the need for explicit positional queries, we employ a pixel-by-pixel decoder.
The decoder, denoted as $f_d$ in Eq.~\ref{eq:total_equation}, employs a simple 5-layer MLP $f_{mlp}^{\circlearrowright^5}$ architecture. The reconstructed output $I$ after decoding can be described in Eq.~\ref{eq:decoder}, where $\oplus$ means element-wise addition.
}
\begin{equation}
    I = f_d(f_{te}(t), f_e(E, I_{rsb})) = f_d(T, \theta) = f_{mlp}^{\circlearrowright^5}(T \oplus \theta).
    \label{eq:decoder}
\end{equation}
\noindent\textbf{Loss Function:}
\label{sec:Loss Function}
Inspired by EVDI~\citep{zhang2022unifying}, we formulate the relationship between RS blur frames and RS sharp frames.
Given a sequence of RS sharp frames generated from the decoder, the input RS blur frame $I_{rsb} = \frac{1}{M} \sum_{i=1}^{M}{(\hat{I}^i_{rss})}$,
where $M$ represents the length of the RS frame sequence.
In this way, we can formulate the blur frame guidance integral loss between the reconstructed RS blur frame and the original RS blur frame as $\mathcal{L}_{b}=\mathcal{L}_{c}(\hat{I}_{rsb}, I_{rsb})$,
where $\mathcal{L}_{c}$ indicates \textit{Charbonnier loss}~\citep{lai2018fast}.
Apart from RS blur image-guided integral loss $\mathcal{L}_{b}$, we incorporate a reconstruction loss $\mathcal{L}_{re}$ to supervise the reconstructed GS sharp frames.
Our total loss function is detailed in Eq.~\ref{eq:total-loss-function}, where $\lambda_{b}$,$\lambda_{re}$ denote the weights of each part.
\begin{equation}
    \mathcal{L} = \lambda_{b}\mathcal{L}_{b} + \lambda_{re}\mathcal{L}_{re}   = \lambda_{b}  \mathcal{L}_{c}(\hat{I}_{rsb}, I_{rsb}) + \lambda_{re} \frac{1}{N} \sum_{k=1}^{N}\mathcal{L}_{c}(\hat{I}^k_{gss}, I^k_{gss}).
\label{eq:total-loss-function}
\end{equation}

\begin{table}[t!]
\caption{ Comparison of properties across different datasets. \label{tab:different-dataset}}
\setlength{\tabcolsep}{0.035\linewidth}{
\resizebox{\linewidth}{!}{
\begin{tabular}{l|cc|rccr}
\toprule
Dataset    & 
Real-world & 
Has GT &
\makecell[c]{Rolling\\Time ($ms$)}   & 
\makecell[c]{Support\\VFI}   & 
\makecell[c]{Blur} &
\makecell[c]{Exposure\\Time ($ms$)} \\
\midrule
Fastec-\textit{Orig}~\citep{liu2020deep}             & \ding{55}  & \ding{51}  & $100.0$               & \ding{55} & Sharp & $0.4$           \\
Gev-\textit{Orig}~\citep{zhou2022evunroll}           & \ding{55}  & \ding{51}  & $31.5$;~$63.1$        & \ding{55} & Blur  & $12.8$;~$25.4$            \\
Fastec-\textit{Blur}                                 & \ding{55}  & \ding{51}  & $108.3$               & \ding{51} & Blur  & $108.3$    \\
Gev-\textit{Blur}                                    & \ding{55}  & \ding{51}  & $45.6$                & \ding{51} & Blur  & $45.6$     \\
\hline
EvUnRoll-Real~\citep{zhou2022evunroll}               & \ding{51}  & \ding{55}  & -                     & \ding{55} & Sharp & -           \\
\bottomrule
\end{tabular}
}
}
\end{table}
\begin{figure*}[t]
\centering
\includegraphics[width=\linewidth]{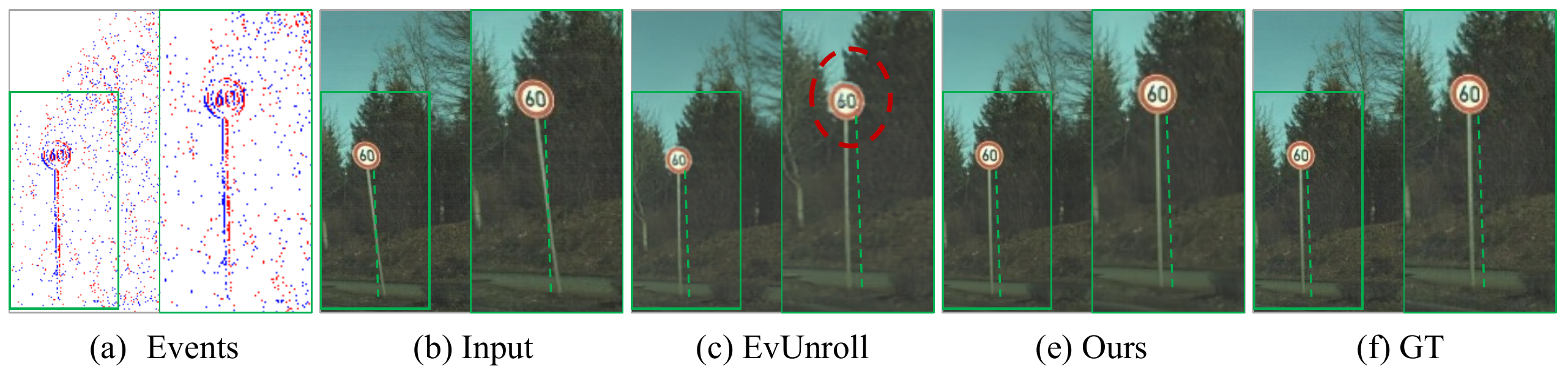}
\caption{Visual results for \textbf{RS correction} on Fastec-\textit{Orig}~\citep{liu2020deep} dataset.\label{fig:Fastec-Sharp-Vis}}
\centering
\end{figure*}
\begin{table}[t]
\centering
{
\caption{Quantitative comparison for \textbf{RS correction} on Fastec-\textit{Orig} dataset~\citep{liu2020deep}.
\label{tab:fastec_simulated_dataset}
}
\resizebox{\linewidth}{!}{
\setlength{\tabcolsep}{0.035\linewidth}{
\begin{tabular}{l|cc|r|clc}
\toprule
Method                                  & Frames    & Event         & \multicolumn{1}{c}{Params($M$)~$\downarrow$}  & PSNR~$\uparrow$   &\multicolumn{1}{c}{SSIM~$\uparrow$}    & LPIPS~$\downarrow$  \\
\midrule
DSUN~\citep{liu2020deep}                & 2         &  \ding{55}    &    3.91~~~                                    & 26.52             & 0.79                                  & 0.122                 \\
RSCD~\citep{zhong2021towards}           & 3         &  \ding{55}    &    7.32~~~                                    & 24.84             & 0.78                                  & 0.107                      \\
SUNet~\cite{fan2021sunet}               & 2         &  \ding{55}    &    11.91~~~                                   & 28.34             & 0.84                                  & -                      \\
ESTRNN~\cite{zhong2020efficient}        & 5         &  \ding{55}    &    2.47~~~                                    & 27.41             & 0.84                                  & 0.189                      \\
JCD~\cite{zhong2021towards}             & 3         &  \ding{55}    &    7.16~~~                                    & 24.84             & 0.78                                  & 0.107             \\
RSSR~\cite{fan2021inverting}            & 2         &  \ding{55}    &    26.04~~~                                   & 21.23             & 0.78                                  & 0.166 \\
CVR~\cite{fan2022context}               & 2         &  \ding{55}    &    42.69~~~                                   & 28.72             & 0.85                                  & 0.111 \\
SelfUnroll-S~\cite{wang2023self}        & 1         &  \ding{51}    &     -~~~~~                                    & 31.85             & 0.89                                  & 0.072        \\
SelfUnroll-M~\cite{wang2023self}        & 2         &  \ding{51}    &     -~~~~~                                    & 32.34             & 0.90                                  & 0.072  \\
EvUnroll~\cite{zhou2022evunroll}        & 1         &  \ding{51}    &    20.83~~~                                   & 31.32             & 0.88                                  & 0.084                      \\
EvShutter~\cite{erbach2023evshutter}    & 1         &  \ding{51}    &     -~~~~~                                    & \underline{32.41}             & \underline{0.91}                                  & \underline{0.061}                     \\
UniINR(Ours)                            & 1         & \ding{51}     &   \textbf{0.38}~~~                            & \textbf{33.91}    & \textbf{0.9234}                       & \textbf{0.049}   \\
\bottomrule
\end{tabular}
}}
}
\end{table}

\section{Experiments}
\noindent\textbf{Implementation Details:}
We utilize the Adam optimizer~\citep{kingma2014adam} for all experiments, with learning rates of $1e-4$ for simulation datasets~\citep{zhou2022evunroll,liu2020deep}.
Using two NVIDIA RTX A5000 GPU cards, we train our framework across 400 epochs with a batch size of two.
In practice, we use the mixed precision~\citep{micikevicius2017mixed} tool provided by PyTorch~\citep{paszke2017automatic}
, which can speed up training and reduce memory usage.
PSNR, SSIM~\citep{wang2004image}, and LPIPS~\cite{zhang2018unreasonable} are used to evaluate the reconstructed results.

\begin{figure*}[t!]
\centering
\includegraphics[width=\linewidth]{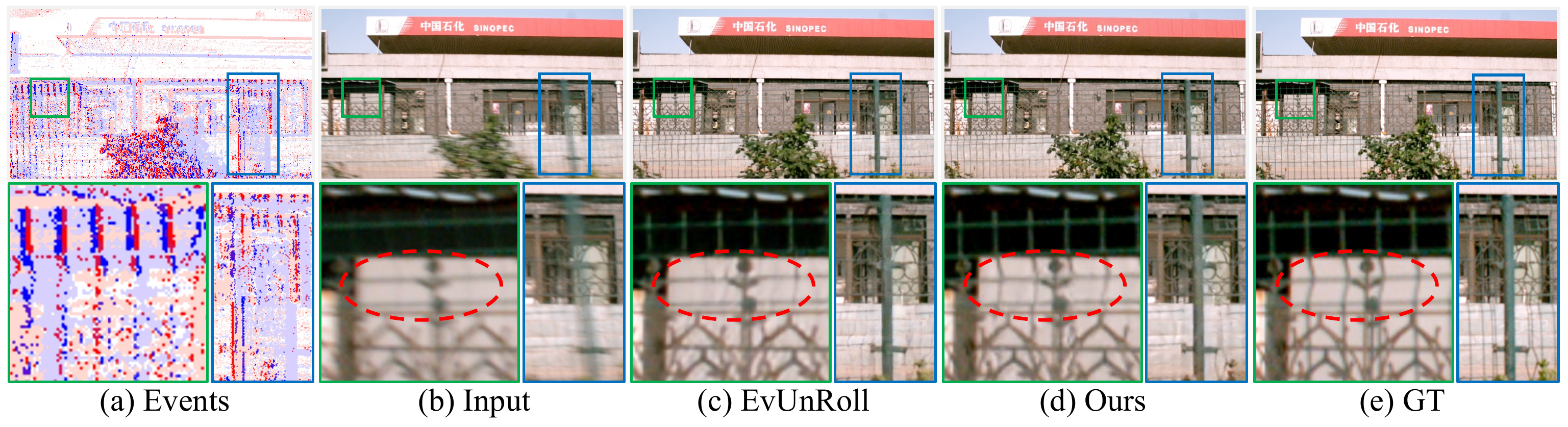}
\caption{Visual results for \textbf{RS correction+Deblur} on Gev-\textit{Orig}~\citep{zhou2022evunroll} dataset.\label{fig:17-Gev-Orig-Release}}
\centering
\end{figure*}

\begin{wrapfigure}{lt}{190pt}
\centering
\caption{Quantitative comparison for \textbf{RS correction + deblurring} on Gev-\textit{Orig} dataset~\citep{zhou2022evunroll}. 
The numerical results of DSUN, JCD, and EvUnRoll are provided by \citep{zhou2022evunroll}.
\label{tab:evunroll_simulated_dataset}}
\resizebox{\linewidth}{!}{
\setlength{\tabcolsep}{0.039\linewidth}{
\begin{tabular}{l|clc}
\toprule
Method                                  & PSNR~$\uparrow$   &\multicolumn{1}{c}{SSIM~$\uparrow$}    & LPIPS~$\downarrow$  \\
\midrule
DSUN~\citep{liu2020deep}                & 23.10             & 0.70                                  & 0.166 \\
JCD~\citep{zhong2021towards}            & 24.90             & 0.82                                  & 0.105  \\
EvUnRoll~\citep{zhou2022evunroll}       & \underline{30.14} & \underline{0.91}                      & \underline{0.061}  \\
NIRE~\citep{zhang2023neural}            & 29.86             & \underline{0.91}                      & - \\
UniINR(Ours)                            & \textbf{31.47}    & \textbf{0.9327}                       &  \textbf{0.038}     \\
\bottomrule
\end{tabular}
}}
\end{wrapfigure}

\begin{figure*}[t]
\centering
\includegraphics[width=\linewidth]{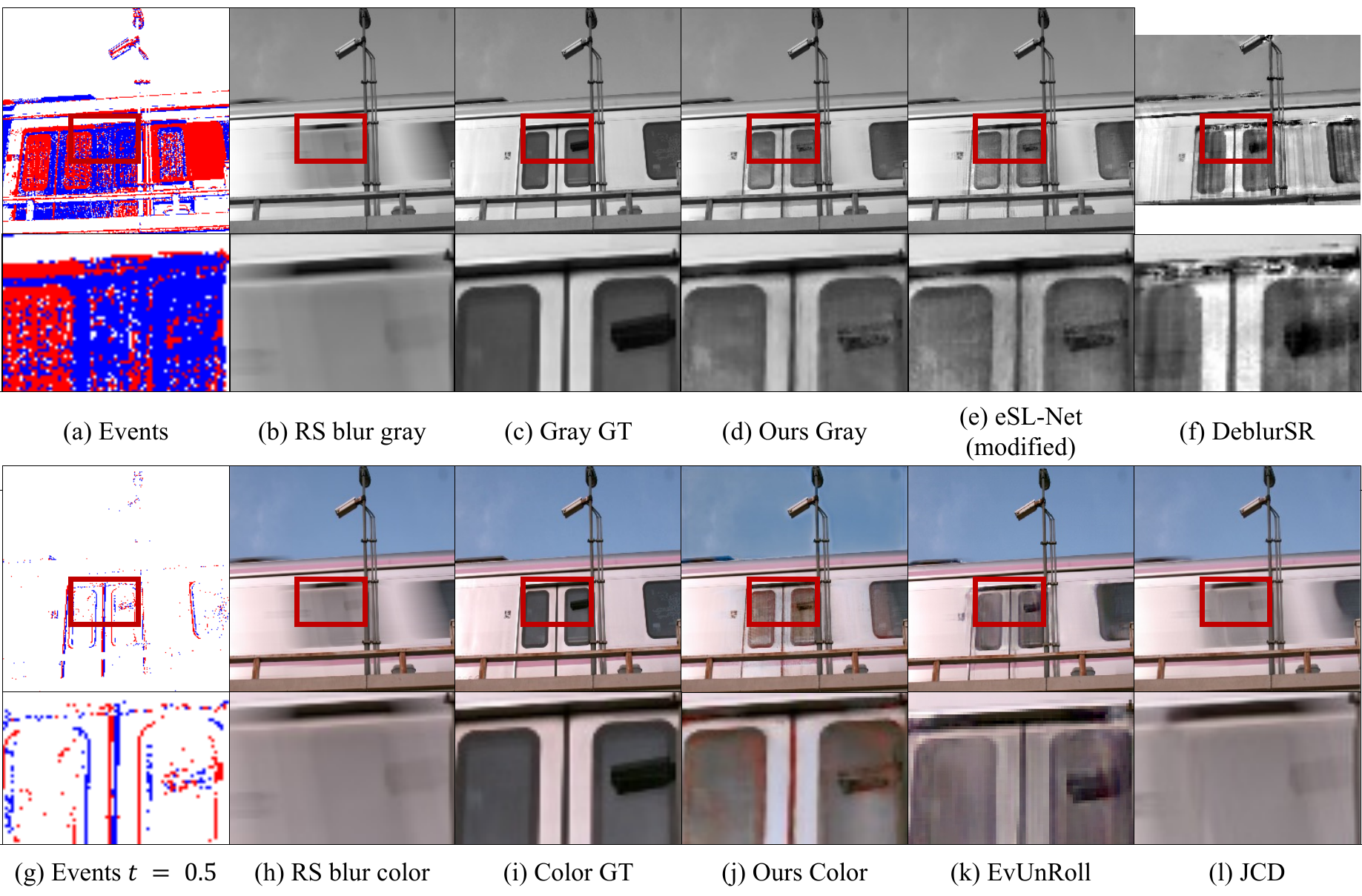}
\caption{Visual Comparisons on \textbf{RS correction + deblurring} on Gev-\textit{Blur}~\citep{zhou2022evunroll} dataset.
The image resolution of DeblurSR~\citep{song2023deblursr} is $180\times 240$.}
\label{fig:gev-rs-blur-vfi}
\centering
\end{figure*}

\begin{figure*}[t]
\centering
\includegraphics[width=\linewidth]{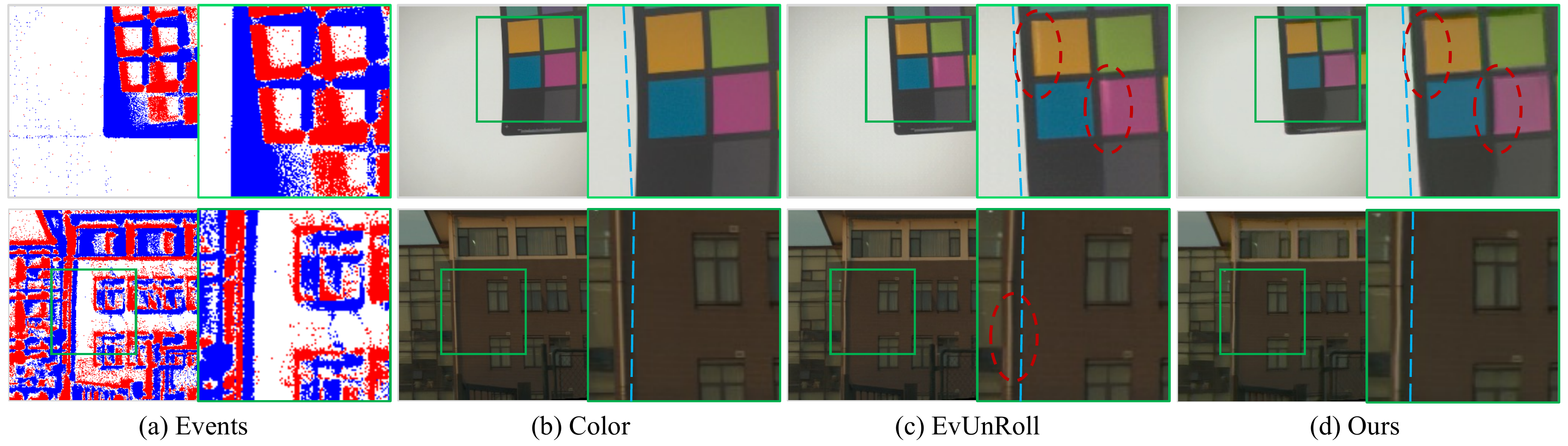}
\caption{ Visualization results in a real-world dataset~\citep{zhou2022evunroll}. (a) are the events visualization. (b) are the input RS frames with clear distortions. (c) are the outputs of EvUnRoll. (d) are the outputs of our method. The red circle in (c) has color/edge distortions.}
\label{fig:12-RealWorld-Release-mini}
\end{figure*}

\noindent \textbf{Setting:}
To verify the performance on three tasks—RS correction, deblurring, and VFI—our experimental setting utilizes a layered task strategy to validate the effectiveness.
We executed experiments across three configurations:
\textbf{(1)} RS correction (Tab.~\ref{tab:fastec_simulated_dataset}, Fig.~\ref{fig:Fastec-Sharp-Vis}),
\textbf{(2)} RS correction + deblurring (Tab.~\ref{tab:evunroll_simulated_dataset},~\ref{tab:rsc_deblur_our_blur_dataset}, Fig.~\ref{fig:17-Gev-Orig-Release},~\ref{fig:gev-rs-blur-vfi}), and
\textbf{(3)} RS correction + deblurring + VFI (Tab.~\ref{tab:rsc_deblur_vfi_our_blur_dataset}, Fig.~\ref{fig:gev-rs-blur-vfi}).
This setting allows for a thorough assessment of the performance.
Additionally, to evaluate our method's generalization capabilities, we conducted tests on the real-world dataset (Fig.~\ref{fig:12-RealWorld-Release-mini}).
\noindent \textbf{Datasets:}
In this process, we used a total of five datasets, as shown in Tab.~\ref{tab:different-dataset}.
These datasets include simulation datasets and real-world datasets.
The simulation datasets are simulated based on high-frame-rate video and can support model training.
Among them, Fastec-\textit{Orig}~\citep{liu2020deep} and Gev-\textit{Orig}~\citep{zhou2022evunroll} are the simulation data sets used in the original paper respectively.
Given the limitations of the original datasets in terms of mild blurring and their inability to support VFI training, we generate the Fastec-\textit{Blur} and Gev-\textit{Blur} datasets to better meet the training needs.
Moreover, we utilize a real-world dataset, EvUnRoll-Real ~\cite{zhou2022evunroll},
to facilitate qualitative visual assessments. However, the absence of ground truth in the dataset precludes the provision of quantitative evaluations.
\textit{More explanations of the setting and datasets are in Supp. Mat.}.

\begin{table*}[t]
    \centering
\caption{
Quantitative results for \textbf{RS correction + deblurring} with \textbf{color and gray} frames on Gev-\textit{Blur} Dataset. 
eSL-Net* is a modified version of eSL-Net. \cite{wang2020event}.
\label{tab:rsc_deblur_our_blur_dataset}
}
\resizebox{\linewidth}{!}{
\setlength{\tabcolsep}{0.015\linewidth}{
{
    \begin{tabular}{cl|lc|r|cccc}
    \toprule
                &                         & \multicolumn{2}{c}{Inputs}                &    ~      & \multicolumn{2}{c}{Gev-RS-\textit{Blur}}       & \multicolumn{2}{c}{Fastec-RS-\textit{Blur}}      \\
                & Methods                 & Frame       & Event     & Params($M$)~$\downarrow$ & PSNR~$\uparrow$ & SSIM~$\uparrow$                   & PSNR~$\uparrow$          & SSIM~$\uparrow$     \\
    \midrule
\multirow{3}{*}{$1\times$}
                & JCD                     & $3$ color   & \ding{55} & 7.1659    & 19.42           & 0.6364            & 24.92             & 0.7422              \\
                & EU                      & $1$ color   & \ding{51} & 20.83     & 26.18           & 0.8606            & 29.76             & 0.8693             \\
                & UniINR (Ours)           & $1$ color   & \ding{51} & 0.3792    & \textbf{30.35}  &\textbf{ 0.9714}   & \textbf{33.64}    &\textbf{0.9299}             \\
    \hline
    \midrule
\multirow{2}{*}{$1\times$}
                & eSL-Net*                & $1$ gray    & \ding{51} & 0.1360    & 31.64           & 0.9614            &32.45              & 0.9186             \\
                & UniINR (Ours)           & $1$ gray    & \ding{51} & 0.3790    & \textbf{33.12}  & \textbf{0.9881}   &\textbf{34.62}     & \textbf{0.9390}                   \\
    \bottomrule
    \end{tabular}
    }
}
}
\end{table*}

\begin{table*}[t]
    \centering
\caption{
Quantitative results for \textbf{RS correction + deblurring + VFI} with \textbf{color and gray} frames.
TL refers to TimeLens~\cite{tulyakov2021time}.
EU refers to EvUnroll~\cite{zhou2022evunroll}.
\label{tab:rsc_deblur_vfi_our_blur_dataset}}
\resizebox{\linewidth}{!}{
\setlength{\tabcolsep}{0.015\linewidth}{
{
    \begin{tabular}{cl|lc|r|cccc}
    \toprule
                &                         & \multicolumn{2}{c}{Inputs}                &    ~      & \multicolumn{2}{c}{Gev-RS-\textit{Blur}}       & \multicolumn{2}{c}{Fastec-RS-\textit{Blur}}      \\
                & Methods                 & Frame       & Event     & Params($M$)~$\downarrow$ & PSNR~$\uparrow$ & SSIM~$\uparrow$                   & PSNR~$\uparrow$          & SSIM~$\uparrow$     \\

    \midrule
\multirow{2}{*}{$3\times$} 
                & EU + TL                 &$2$ color    & \ding{51} & 93.03     &21.86          & 0.7057        &  24.81        & 0.7179          \\
                & UniINR (Ours)           &$1$ color    & \ding{51} & 0.3792    &\textbf{28.36} &\textbf{0.9348}&\textbf{ 32.72}&	\textbf{0.9147}        \\
    \hline
\multirow{2}{*}{$5\times$}
                & EU + TL                 &$2$ color    & \ding{51} & 93.03     & 21.59         & 0.6964        & 24.46         & 0.7140            \\
                & UniINR (Ours)           &$1$ color    & \ding{51} & 0.3792    &\textbf{28.41} &\textbf{0.9062}&  \textbf{32.13}&	\textbf{0.9053}          \\
    \hline
\multirow{2}{*}{$9\times$}
                & EU + TL                 &$2$ color    & \ding{51} & 93.03     & 21.24         & 0.6869        & 23.99         & 0.7029          \\
                & UniINR (Ours)           &$1$ color    & \ding{51} & 0.3792    &\textbf{27.21} &\textbf{0.8869}& \textbf{29.31}&	\textbf{0.8590}          \\
    \hline
    \midrule
\multirow{2}{*}{$3\times$}& DeblurSR      &$1$ gray     & \ding{51} & 21.2954   &  17.64        & 0.554         & 21.17         & 0.5816    \\
                & UniINR (Ours)           &$1$ gray     & \ding{51} & 0.3790    &\textbf{31.11} &\textbf{0.9738}&\textbf{33.23} &	\textbf{0.9210}         \\
    \hline
\multirow{2}{*}{$5\times$}& DeblurSR      &$1$ gray     & \ding{51} & 21.2954   & 18.35         & 0.6107        & 22.86         & 0.6562           \\
                & UniINR (Ours)           &$1$ gray     & \ding{51} & 0.3790    &\textbf{30.84} &\textbf{0.9673}&\textbf{32.82} &	\textbf{0.9147}         \\
    \hline
\multirow{2}{*}{$9\times$}& DeblurSR      &$1$ gray     & \ding{51} & 21.2954   & 18.86         & 0.6502        & 23.96         & 0.7049            \\
                & UniINR (Ours)                                                 &$1$ gray & \ding{51}    & 0.3790    &\textbf{30.54} &\textbf{0.9579}& \textbf{32.21}&	\textbf{0.9051}        \\
    \bottomrule
    \end{tabular}
    }
}
}
\end{table*}


\noindent\textbf{Comparison Experiments:}
\noindent\textbf{\textit{(1) RS Correction:}} We conducted a comparison of RS correction on the Fastec-\textit{Orig} dataset~\citep{liu2020deep}, comparing frame-based methods~\citep{liu2020deep,zhong2021towards,fan2021sunet,zhong2020efficient,zhong2021towards,fan2021inverting,fan2022context} and event-based methods~\citep{zhou2022evunroll,wang2023self,erbach2023evshutter} in Tab.~\ref{tab:fastec_simulated_dataset}.
The results show that our model size is only $1/10$ to $1/110$ of the comparative models, yet it surpasses existing models in all evaluation metrics.
Notably, our model exceeds the current state-of-the-art (SOTA) by \textbf{1.6}$dB$ in PSNR.
Moreover, in Fig.~\ref{fig:Fastec-Sharp-Vis}, we display the visual results of RS correction.
Compared to EvUnRoll, our method significantly improves deformation correction and achieves clearer edges ('60' in the red circle).
\noindent\textbf{\textit{(2) RS Correction + Deblurring:}}
We compare the combined task of RS correction and deblurring.
In Tab.~\ref{tab:evunroll_simulated_dataset}, we compared two frame-based methods~\cite{liu2020deep,zhong2021towards} and two event-based methods~\cite{zhou2022evunroll,zhang2023neural} on the Gev-\textit{Orig} dataset, which accounts for slight blur (with exposure times of $12 ms$ and $25 ms$) as shown in Tab.~\ref{tab:different-dataset}.
The results demonstrate the advantage of UniINR, surpassing the SOTA method, EvUnRoll~\cite{zhou2022evunroll}, by \textbf{1.33}$dB$.
Furthermore, we explored our method's performance under greater blur degrees, as detailed in Tab.~\ref{tab:rsc_deblur_our_blur_dataset}.
The Gev-\textit{Blur} dataset, with a $45.6 ms$ exposure, displays nearly twice the blur of Gev-\textit{Orig} ($25.4 ms$), and Fastec-\textit{Blur} extends exposure to $108 ms$.
Our approach consistently outperformed, particularly against the SOTA method, EvUnroll, which suffers from error accumulation during its two-stage deblurring and RS correction process.
This issue is more evident with higher blur degrees, as shown in Tab.~\ref{tab:rsc_deblur_our_blur_dataset}.
Our method maintained stability and effectively avoided accumulated errors; notably, even as the blur intensity doubled, our PSNR experienced a minimal decline, moving from $31.47 dB$ to $30.35 dB$, a decrease of only $1.12 dB$. This stands in stark contrast to EvUnRoll, which saw its PSNR drop significantly from $30.14 dB$ to $26.18 dB$, a substantial fall of $3.96 dB$.
Visual results also demonstrate our method's strength, especially in Fig.~\ref{fig:17-Gev-Orig-Release}, where we successfully recover edges lost to blur.
This capability is further evident in Fig.~\ref{fig:gev-rs-blur-vfi}, where our method restores sharp edges in fast-moving trains under higher blur degrees, unlike outputs of EvUnRoll and JCD.
Our one-stage strategy for deblurring and RS correction avoids error accumulation, effectively handling complex motion and showing better results.
\noindent\textbf{\textit{(3) RS Correction + Deblurring + VFI:}}
\begin{wraptable}{lt}{180pt}
\centering
\caption{Ablation for position embedding.}\label{tab:ablation_encoding}
\resizebox{\linewidth}{!}{
\setlength{\tabcolsep}{0.039\linewidth}{
\begin{tabular}{ccrr}
\toprule
                              & Position Embedding & PSNR           & SSIM   \\
\midrule
\multirow{2}{*}{$1\times$}    & Sinusoid           & 32.46          &	0.9851         \\
                              & Learning           & \textbf{33.12} & \textbf{0.9881} \\
\midrule
\multirow{2}{*}{$3\times$}    & Sinusoid           & 30.83          & 0.9723         \\
                              & Learning           & \textbf{31.11} & \textbf{0.9738} \\
\midrule
\multirow{2}{*}{$5\times$}    & Sinusoid           & 30.70          & \textbf{0.9678} \\
                              & Learning           & \textbf{30.84} & 0.9673 \\
\midrule
\multirow{2}{*}{$9\times$}    & Sinusoid           & 30.51          & 0.9560  \\
                              & Learning           & \textbf{30.54} & \textbf{0.9579} \\
\midrule
                             ~&                   ~& \textbf{+1.11} & \textbf{+0.0059} \\
\bottomrule
\end{tabular}
}}
\end{wraptable}

We undertook detailed comparative experiments across three tasks on Gev-\textit{Blur} and Fastec-\textit{Blur} datasets, assessing our method's performance showcased in Tab.~\ref{tab:rsc_deblur_vfi_our_blur_dataset}.
Our method exceeds DeblurSR~\citep{song2023deblursr} and EvUnroll~\cite{zhou2022evunroll}+TimeLens~\cite{tulyakov2021time} by up to \textbf{13.47dB} and \textbf{8.49dB}.
DeblurSR~\citep{song2023deblursr} falls short by focusing only on deblurring and VFI, overlooking RS distortion's effects.
Similarly, EvUnroll~\cite{zhou2022evunroll}+TimeLens~\cite{tulyakov2021time} underperforms due to accumulated errors in the cascading network, as evidenced in Fig.~\ref{fig:inputs_and_outputs}(h).
The visual results presented in Fig.~\ref{fig:gev-rs-blur-vfi} highlight the effectiveness of our method for both grayscale and color inputs, successfully generating sharp frames free from RS distortion even in challenging scenarios like a fast-moving train.
In contrast, eSL-Net and EvUnroll's outputs contain noticeable artifacts, especially around the train door as marked in the red region of Fig.~\ref{fig:gev-rs-blur-vfi}.
Moreover, DebluSR~\citep{song2023deblursr} results show significant artifacts, further emphasizing the superiority of our method in handling multiple tasks simultaneously.
\textit{For more visualization results, please refer to demo videos in Supp. Mat..}
\noindent\textbf{\textit{Real-world Dataset:}}
Fig. \ref{fig:12-RealWorld-Release-mini} shows real-world results. The input frame exhibits rolling shutter distortions, such as curved palette edges. In contrast, events show global shutter traits. Both our method and EvUnRoll correct these distortions effectively.
Due to the lack of ground truth, quantitative analysis is not possible.
Notably, our method avoids artifacts and errors, outperforming EvUnRoll in palette scenarios and building.
\textit{More discussion please refer to the Supp. Mat..}

\begin{wrapfigure}{lt}{200pt}
    \centering
    \includegraphics[width=\linewidth]{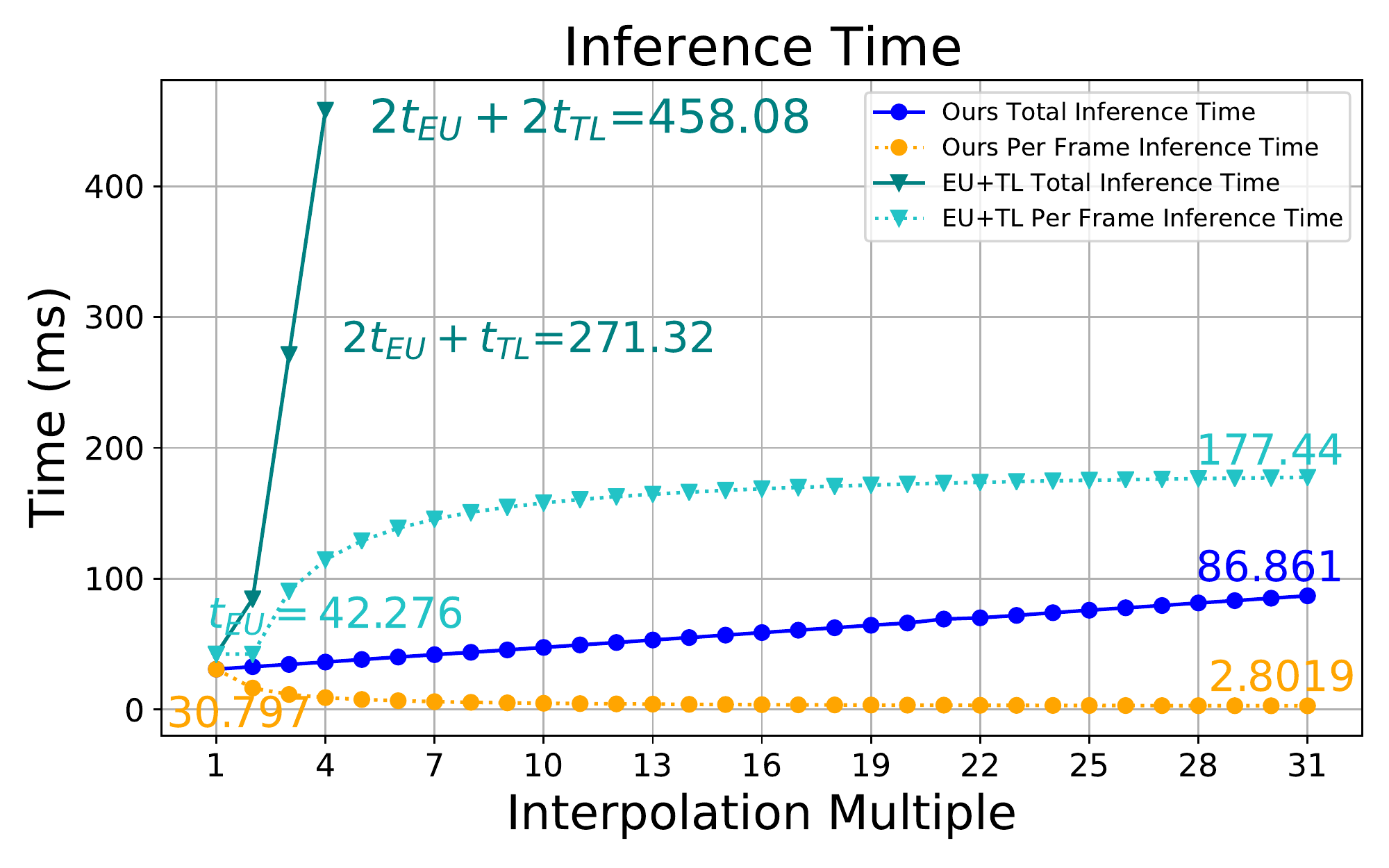}
    \caption{
Comparison of inference time of our method with EvUnroll + TimeLens. $t_{EU}$ and $t_{TL}$ represent the respective inference times of EvUnRoll and TimeLens. The axes represent VFI multiples (\(1\times\) to \(31\times\)) and time.
$2T_{EU}$ and $2 t_{TL}$ means calling EvUnRoll twice and TimeLens twice.
}
    \label{fig:inference-time}
    \centering
\end{wrapfigure}

\noindent\textbf{Ablation and Analytical Experiments: \label{sec:analytical_studies}}
\noindent\textit{\textbf{Importance of Exposure Time Embedding:}}
We conduct the experiments to evaluate the impact of learning-based position embedding, with a comparative analysis to sinusoid position embedding~\citep{vaswani2017attention}.
As indicated in Tab.~\ref{tab:ablation_encoding}, learning-based position embedding outperforms sinusoid position embedding, with advancements of up to \textbf{1.11dB} on average.
This superior efficacy is attributable to the intrinsic adaptability of the learning-based position embedding.
\noindent\textit{\textbf{Inference Speed:}}
Fig.~\ref{fig:inference-time} shows our method's inference time across $1\times$ to $31\times$ interpolation. The total time rises modestly, e.g., from $30.8~ms$ at $1\times$ to $86.9~ms$ at $31\times$, a $2.8$-fold increase for a $31$-fold interpolation. The average frame time even decreases at higher multiples, reaching $2.8~ms$ at $31\times$.
Compared to EvUnRoll~\citep{zhou2022evunroll} and TimeLens~\citep{tulyakov2021time}, our method is more computationally efficient, requiring only 72\% of EvUnRoll's $42.3~ms$ for RS correction and deblurring.
For $N$-fold frame insertion using EvUnRoll + TimeLens, EvUnRoll is counted twice, and TimeLens $N-2$ times.
This advantage is amplified in high-magnification scenarios, where TimeLens costs $186.76~ms$ per call. Our calculations focus on GPU time, excluding data I/O, further increasing EvUnRoll and TimeLens' time consumption.

\section{Conclusion}
This paper presented a novel approach that simultaneously uses events to guide rolling shutter frame correction, deblur, and interpolation.
Unlike previous network structures that can only address one or two image enhancement tasks, our method incorporated all three tasks concurrently, providing potential for future expansion into areas such as image and video super-resolution and denoising.
Furthermore, our approach demonstrated high efficiency in computational complexity and model size.
Regardless of the number of frames involved in interpolation, our method only requires a single call to the encoder, and the model size is small.

{\subsection*{Acknowledgements:}
This work was supported in part by the National Natural Science Foundation of China (Grant No.92370204),
in part by Guangzhou-HKUST(GZ) Joint Funding Program (Grant No.2023A03J0008),
in part by OPPO Research Fund,
in part by the Education Bureau of Guangzhou Municipality.}

\clearpage
{
\small
\bibliography{egbib}
\bibliographystyle{splncs04}
}
\clearpage
\appendix
\section{Events Generation}
\label{sec:event_generation}
The event stream $E$ consists of a set of event $e=(x,y,t,p)$, where each event is triggered and recorded when the brightness change at pixel $(x,y)$ exceeds a certain threshold $C$.
The time interval between events is denoted as $\Delta t_e$, which is a short period, and the brightness at position $\bm{x} = (x,y)$ is represented as $F(\bm{x},t)$.
The brightness change can be calculated as $\Delta b = log(F(\bm{x},t_e))-log(F(\bm{x},t_e - \Delta t_e))$.
The output signal $p$ is determined by Eq.~\ref{eq:event-generation}.

\begin{equation}
p = \left\{
\begin{array}{cc}
     1, & \Delta b > C  \\
     0, & others \\
     -1,& \Delta b < -C
\end{array}
\right.
\label{eq:event-generation}
\end{equation}

\section{The Encoding Network Structure}
\label{sec:encoding_structure}
\begin{figure*}[h]
\centering
\includegraphics[width=\linewidth]{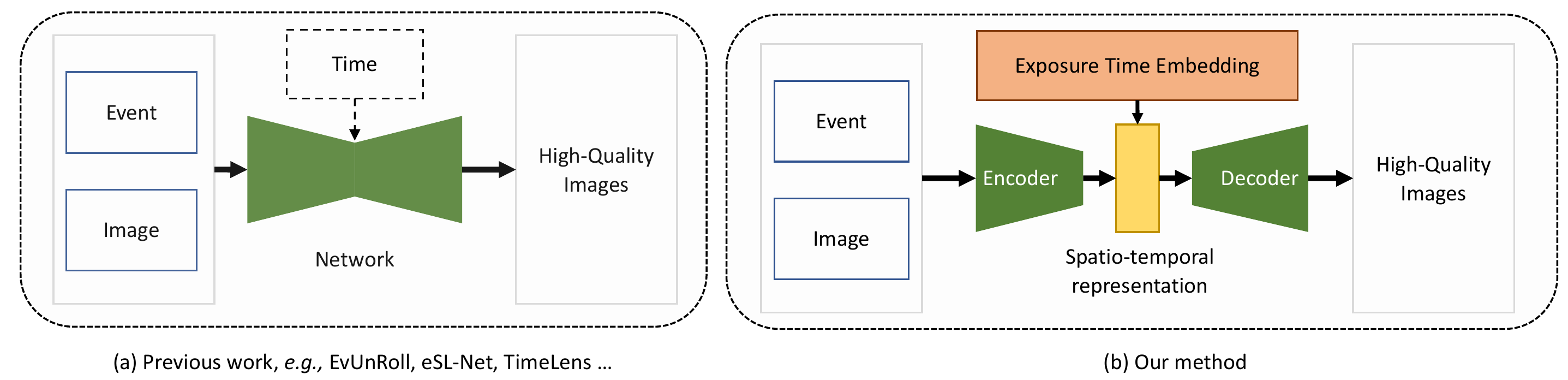}
\caption{Differences between our method and previous methods. In contrast to the previous method, our approach introduces spatio-temporal representation and exposure time embedding. The spatio-temporal representation involves capturing all the spatio-temporal information during the exposure time. Furthermore, specific exposure time information is embedded, which enables the decoder to generate a frame with high-quality.\label{fig:previous-methods-our-methods}}
\centering
\end{figure*}

The encoder serves as the core component of our architecture, drawing inspiration from eSL-Net\citep{wang2020event}. However, we have made modifications to its structure by excluding the decoding segment responsible for upsampling.
Consequently, we retain solely the feature extraction module, as illustrated in Fig.~\ref{fig:previous-methods-our-methods}, and the code of the encoder is shown in Code (Listing. {\color{hollywoodcerise} 1}).

The encoder receives inputs, rolling shutter blur image $I_{rsb}$, and events $E$. The image $I_{rsb}$ has the shape of $H\times W\times 1$ or $H\times W\times 3$, corresponding to grayscale and RGB image, respectively.
Moreover, the event $E$ is transformed into count images~\citep{zheng2023deep}, with the shape of $H\times W\times M$, where $M$ denotes the number of temporal divisions within the event stream.
The encoder produces a high-dimensional tensor as output $\theta$, with the shape of $H\times W\times C$.

The encoder can be decomposed into two constituent components: data preprocessing and spatio-temporal information modeling. During the preprocessing stage, the image undergoes a convolution operation to augment the channel dimensionality, while the event data is transformed into a high-dimensional Tensor using two convolutions followed by a Sigmoid activation. Subsequently, the processed tensors from the image and event data are subjected to the sparse learning module.
Within the sparse learning module, both image features and event features undergo iterative cycles to derive spatio-temporal representations $\theta$.
In contrast to the original approach eSL-Net~\citep{wang2020event}, we aim to incorporate deformable convolutions~\citep{wang2022internimage} into this loop, thereby enhancing the motion estimation and correction capabilities throughout the iterations.

\section{More Explanation and Discussion}
\label{sec:explanation_experiments}

Due to the constraints of the main paper's length, this supplementary section provides additional experimental details and discussions. Specifically, we elaborate on the following 11 aspects to offer readers a more comprehensive understanding of our approach:

\begin{figure*}[h]
\centering
\includegraphics[width=\linewidth]{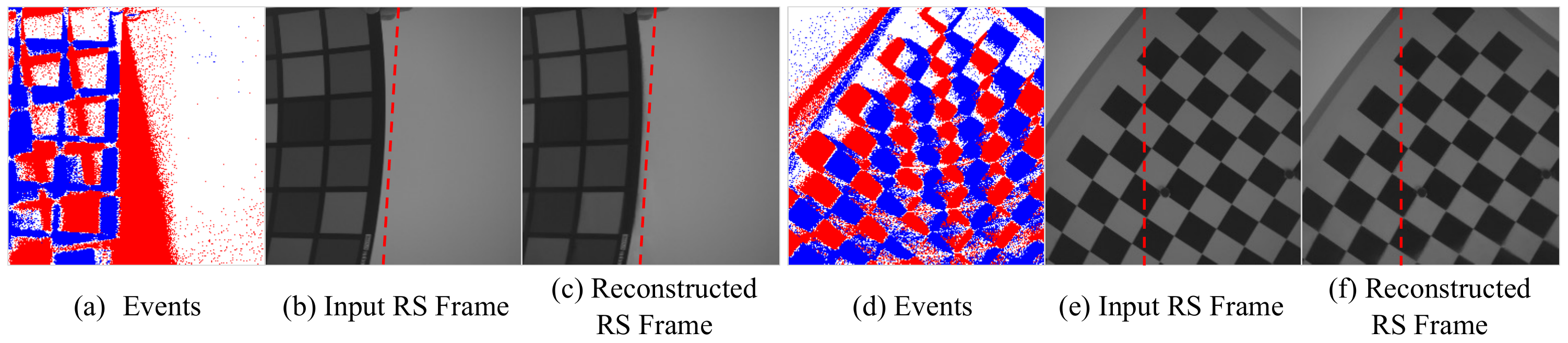}
\caption{Rolling shutter frame reconstruction visualization in the real-world dataset~\citep{zhou2022evunroll}.}
\label{fig:15-blur-reconstruction-mini}
\centering
\end{figure*}

\begin{figure*}[h]
\centering
\includegraphics[width=0.7\linewidth]{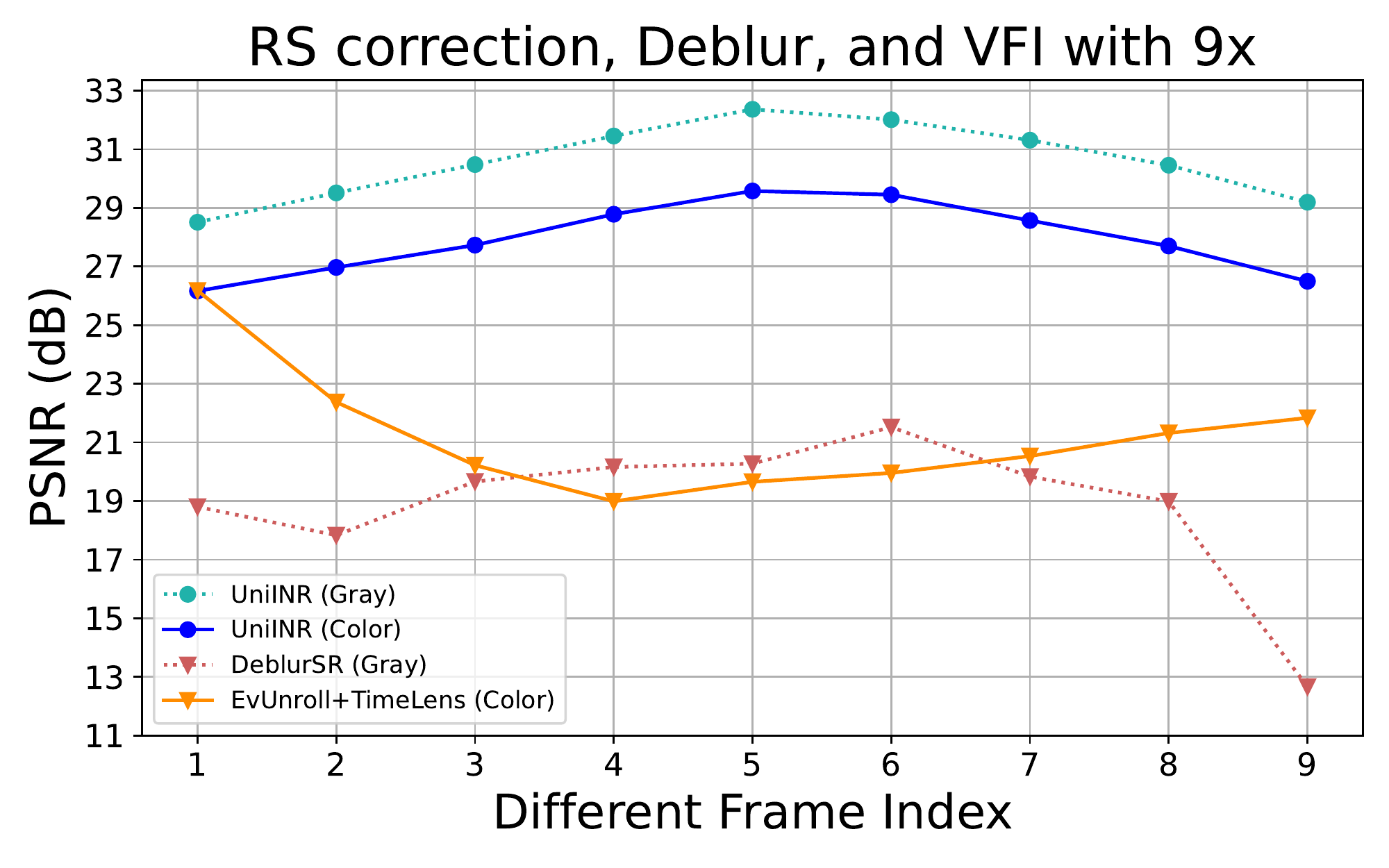}
\caption{{The $x$-axis corresponds to the subscript of the frame interpolation result achieved through 9-fold frame interpolation, while the $y$-axis represents the PSNR value associated with each frame. A higher PSNR value indicates a higher reconstruction quality.\label{fig:psnr-frame-index}}}
\centering
\end{figure*}

\subsection{VFI Performance stability:}
Fig.~\ref{fig:psnr-frame-index} illustrates the PSNR values for each frame obtained through various methods using 9-fold frame interpolation. Our proposed method demonstrates superior performance. Specifically, the intermediate image attains the highest quality, while the reconstruction quality diminishes towards both the beginning and end of the exposure, displaying a symmetrical pattern. To further enhance image quality across the entire frame, future investigations could explore the integration of multi-frame algorithms.

\subsection{RS Blur Image-guided Integral Loss:}
The RS blur image-guided integral Loss enhances PSNR in high interpolation settings (\eg, $9\times$), as shown in Tab.~\ref{tab:loss_function}.
Crucially, we find our model has the generalization ability to reconstruct RS frames in the real-world dataset, as shown in Fig.~\ref{fig:15-blur-reconstruction-mini}.
This underscores our method's adeptness at capturing the temporal intensity dynamics of each pixel for effective generalization in real-world.

\input{Table-3-Ablation-Large}


\subsection{Further Detail on the Dataset:}
\textbf{1) Gev-\textit{Orig} dataset~}~\citep{zhou2022evunroll}
contains original videos shot by GS high-speed cameras with $1280\times720$ resolution at 5700 fps.
However, EvUnroll~\citep{zhou2022evunroll} primarily focuses on RS correction, and provided by EvUnroll Gev-RS dataset does not include RS frames with severe motion blur.
Therefore, we reconstruct RS frames with severe motion blur and events from original videos.
We initially downsample the original videos to DAVIS346 event camera's resolution ($260 \times 346$)~\citep{scheerlinck2019ced}.
Then, we employ the event simulator vid2e~\citep{gehrig2020video} to synthesize events from the resized frames.
We simulate RS blur frames by first generating RS sharp frames as the same RS simulation process of Fastec-RS~\citep{liu2020deep} and then averaging 260 RS sharp frames after gamma correction.
We use the same dataset split as EvUnroll~\citep{zhou2022evunroll}, with 20 videos used for training and 9 videos used for testing.
The total amounts of RS blur frames in Gev-RS~\citep{zhou2022evunroll} dataset are 784 in the training set and 441 testing set.

\noindent\textbf{2) Fastec-\textit{Orig} dataset~}~\citep{liu2020deep}
provides the original frame sequences recorded by the high-speed GS cameras with the resolution of $640\times480$ at 2400 fps.
We use the same settings to resize frame sequences, create events, and RS blurry frames.
Furthermore, we use the same dataset split strategy as Fastec-RS~\citep{liu2020deep}: 56 sequences for training and 20 sequences for testing.
Specifically, this dataset includes 1620 RS blur frames for training and 636 RS blur frames for testing.

\noindent\textbf{3) Real-world dataset} ~\citep{zhou2022evunroll} currently the sole real dataset accessible, comprises four videos. Among these, two capture outdoor scenes, while the other two focus on indoor scenes. Each video pairs rolling shutter frames with events; the events are derived from DVS346. However, given the absence of ground truth in this dataset, it can only provide quantitative visualization results.

\begin{figure*}[h]
\centering
\includegraphics[width=0.99\linewidth]{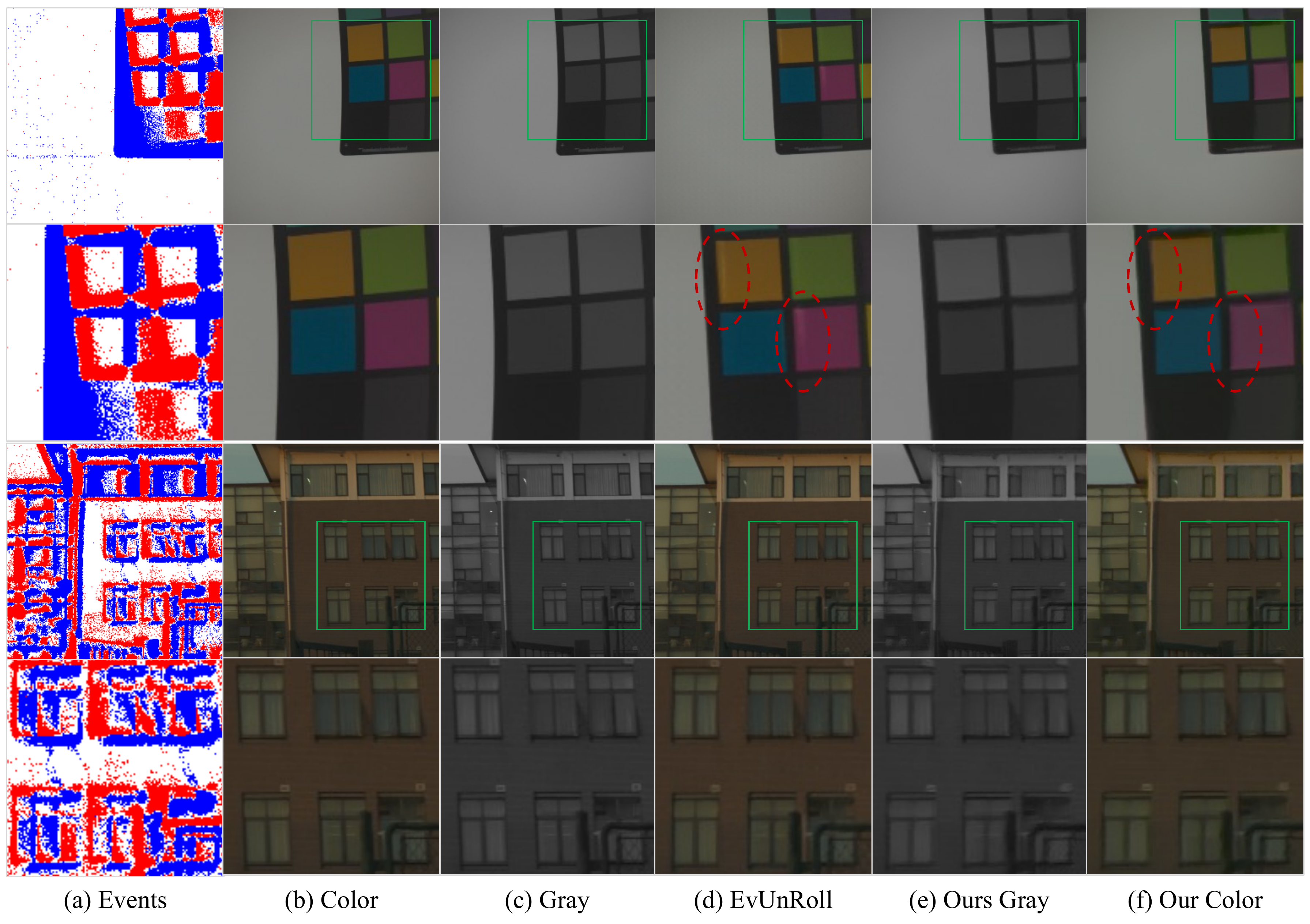}
\caption{\small Visualization results in a real-world dataset ~\citep{zhou2022evunroll}. (a) is the events visualization results. (b) (c) are the input RGB and gray images that have clear rolling shutter distortions. (d) is the output of EvUnRoll. (e) (f) are the outputs of our method.}
\label{fig:12-RealWorld-Release}
\centering
\end{figure*}

\subsection{Bad case analysis: }
Color distortion is due to the input's heavy blur lacking color details. Since events only record intensity (gray) changes, without color information, our method effectively outlines shapes and edges but struggles with color accuracy in this extreme example.

\subsection{Extended Discussion on Inference Speed:}
Fig.~\ref{fig:inference-time} illustrates the inference time of our method with a wide range of interpolation multiples spanning from $1\times$ to $31\times$, including the total inference time and the average inference time per frame.
Importantly, the total inference time increases gradually as the frame interpolation multiple increases. For instance, when going from $1\times$ to $31\times$ frame interpolation, the total inference time only increases from $30.8~ms$ to $86.9~ms$. This signifies a mere $2.8$-fold increase in time despite a $31$-fold increase in the interpolation multiple. Additionally, it is notable that the average inference time per frame decreases with higher frame interpolation multiples. At $31\times$ frame interpolation, the average time per frame is a mere $2.8~ms$.
Our method exhibits distinct advantages over the EvUnRoll~\citep{zhou2022evunroll} and TimeLens~\citep{tulyakov2021time} cascade approaches, particularly in terms of computational efficiency. Specifically, when the focus is solely on RS frame correction and deblurring, the inference time for EvUnRoll is measured at 42.3 ms, while our approach necessitates only 72\% of that time.
This computational advantage becomes even more pronounced during high-magnification frame interpolation.
For instance, in scenarios requiring $N$-times interpolation, the cascading strategy calls for two invocations of EvUnRoll and $(N-2)$ of TimeLens, with the latter having a time cost of $(t_{TL} = 186.76~ms)$.
Consequently, our method offers a significant advantage in high-magnification frame interpolation scenarios. It is crucial to note that our inference time calculations are restricted to GPU-based computations, intentionally omitting the time required for data loading and storage. In practical applications, the EvUnRoll and TimeLens cascade introduces additional disk I/O overhead, thereby further exacerbating its time consumption.

\subsection{Additional Insights into the Real-World Dataset:}
The visualization results for the real-world dataset can be seen in Fig. \ref{fig:12-RealWorld-Release}. The input frame, which displays a rolling shutter pattern, is characterized by clear distortions in dynamic scenes.
For example, the palette's edges are curved, and the building windows tilt.
In contrast, events display global shutter characteristics, as evidenced by the lack of distorted edges in the event visualizations.
Both our method and EvUnRoll effectively correct the rolling shutter distortion, whether it's the distortion of the palette's edge or the deformation of building windows. However, due to the absence of ground truth, quantitative analysis remains unattainable. It's worth noting that while EvUnRoll exhibits some artifacts in the palette scenarios, our method remains artifact-free. By concurrently addressing RSC, Deblur, and VFI, our method avoids accumulating errors, leading to a more artifact-free outcome.

\subsection{More Operations in Exposure Time Embedding}
We perform more experiments on Gev-RS dataset ~\citep{zhou2022evunroll} to validate the effect of element-wise addition, multiplication.
The quantitative result is shown in Tab. \ref{tab:more_operation_in_EET}, and we find that concatenation and multiplication have higher PSNR than element-wise addition.

\begin{table}[t]
\centering
\small{
\caption{More operation studies for exposure time embedding. (Gray blur frame as inputs in 1, 3, 5, 9 times frame interpolation).}
\label{tab:more_operation_in_EET}
\centering
\begin{tabular}{cccc}
\toprule
&Time Embedding Type & PSNR  & SSIM   \\
\midrule
            & Add                          & \textit{33.12}          & \textbf{0.9881} \\
$1 \times$  & Multiplication               & \textbf{33.15}          & 0.9757 \\
            & Concat                       & \textbf{33.15}          & \textit{0.9876}         \\
\hline
            & Add                          & \textit{31.11}          & \textbf{0.9738} \\
$3 \times$  & Multiplication               & 31.10                   & 0.9635  \\
            & Concat                       & \textbf{31.14}          & \textit{0.9710}         \\
\hline
            & Add                          & 30.84                  & \textbf{0.9673} \\
$5 \times$  & Multiplication               & \textbf{30.96}         & \textit{0.9684}  \\
            & Concat                       & \textit{30.89}         & 0.9632         \\
\hline
            & Add                          & 30.54          & \textit{0.9579} \\
$9 \times$  & Multiplication               & \textit{30.74} & \textbf{0.9592} \\
            & Concat                       & \textbf{30.77} & 0.9538         \\
\bottomrule
\end{tabular}
}
\end{table}

\begin{figure*}[t]
\centering
\includegraphics[width=\linewidth]{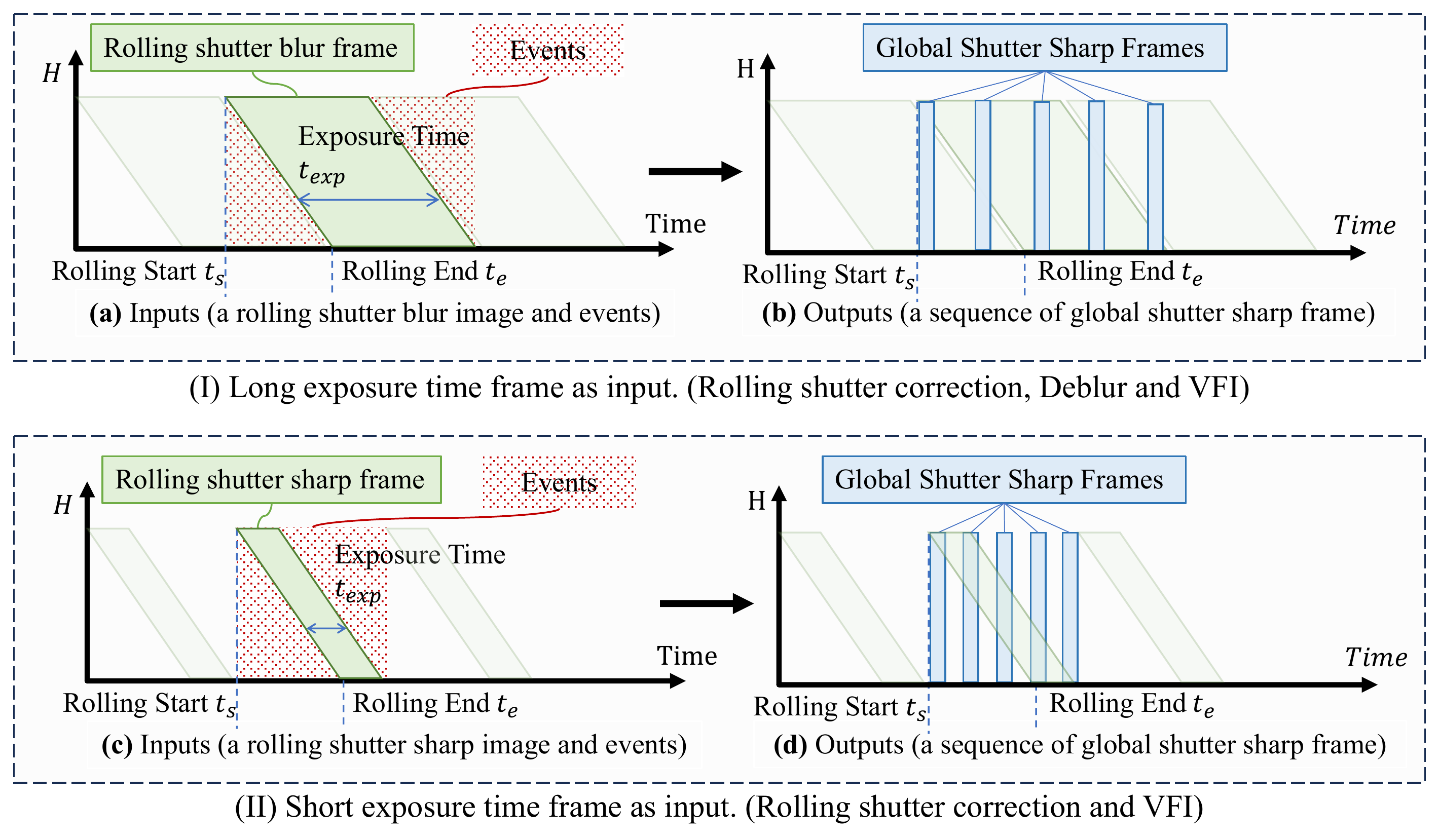}
\caption{The schematic diagram elucidates the methodologies for correcting and interpolating rolling shutter (RS) frames under varying exposure durations. Subfigure (I) delineates the procedure for long-exposure RS frames, where the presence of blur is a significant factor to be addressed. In contrast, Subfigure (II) outlines the approach for short-exposure RS frames, thereby eliminating the necessity for deblurring.}
\label{fig:long-short-exposure-rs-frame}
\centering
\end{figure*}

\begin{figure*}[h]
\centering
\includegraphics[width=\linewidth]{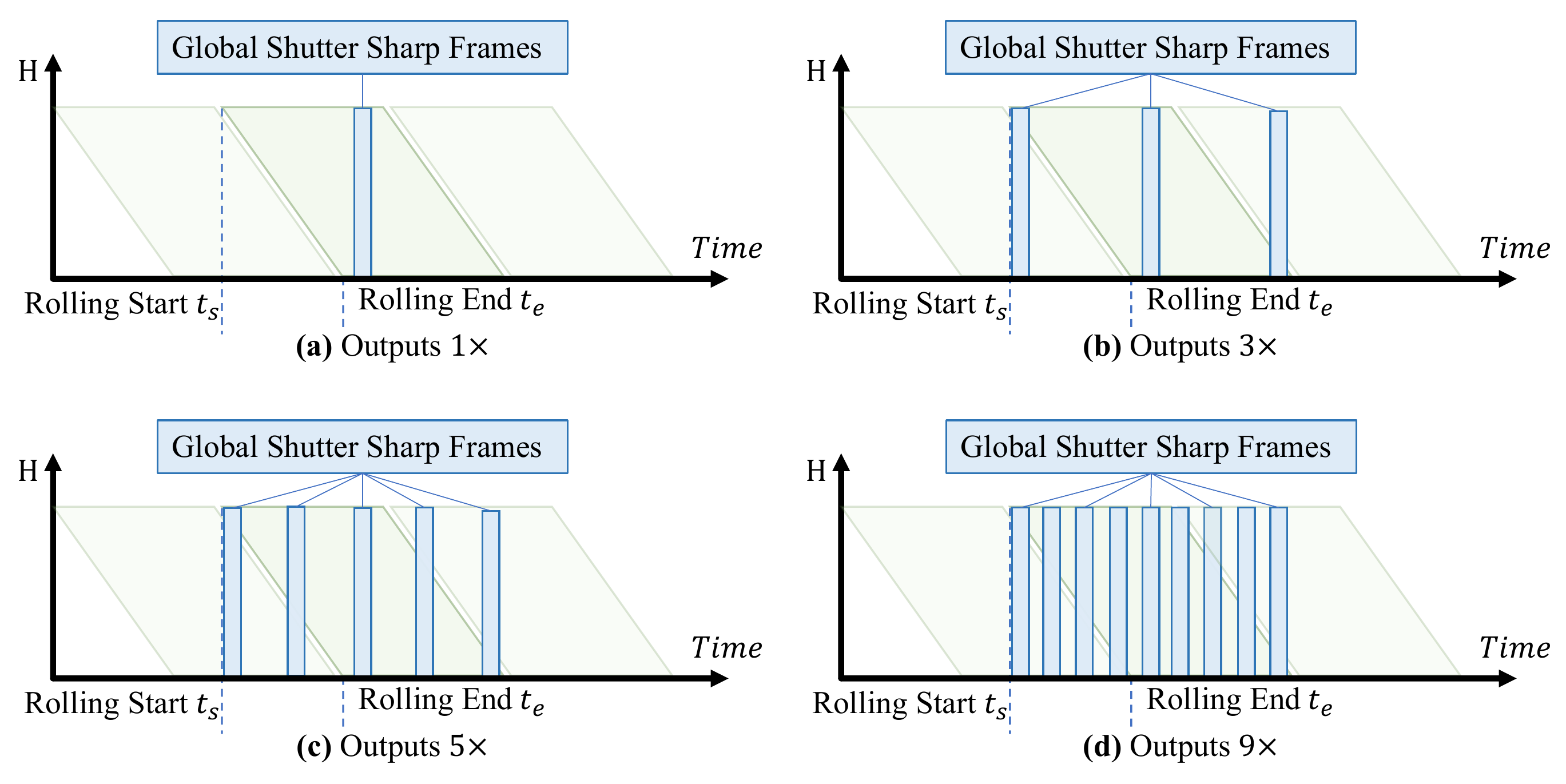}
\caption{Schematic diagram of frame insertion at different magnifications}
\label{fig:different-magnification-interpolation}
\centering
\end{figure*}

\subsection{Analysis of PSNR and SSIM Across Different Interpolation Multiples:}
In Tab.~\ref{tab:rsc_deblur_vfi_our_blur_dataset} of the main manuscript, we observe an intriguing discrepancy in the PSNR and SSIM metrics for $3 \times$ and $5 \times$ color frame interpolations, registering values of 28.36 and 0.9062, and 28.41 and  0.9062, respectively. Contrary to conventional wisdom, which posits that an increase in frame rate interpolation should correspondingly degrade PSNR and SSIM metrics when the model architecture remains constant, our findings deviate from this expectation. We attribute this anomaly to the network's varying predictive accuracy across the temporal spectrum. Specifically, edge frames pose a greater challenge for the network compared to those situated centrally. As illustrated in Fig.~\ref{fig:psnr-frame-index}, for a $3 \times$ frame insertion, the terminal global shutter sharp frames contribute to $2/3$ of the overall weight. Conversely, for a $5 \times$ frame insertion, the terminal frames account for only $2/5$ of the weight.

\subsection{Detailed Comparisons with Other Methods:}

\begin{table}[h]
\centering
\caption{Comparison of our method with prior works. \label{tab:methods_comparison_table}}
\small
\resizebox{1\linewidth}{!}{
\setlength{\tabcolsep}{0.025\linewidth}{
\begin{tabular}{lrcccccc}
\toprule
Methods            & Publication & Frames & Color     & Events    & Deblur          & RS Correction  & VFI \\
\midrule
JCD                & CVPR 2021 & 3      & \ding{51} & \ding{55} & \ding{51}       & \ding{51}           & \ding{55}                    \\
eSL-Net            & ECCV 2020 & 1      & \ding{55} & \ding{51} & \ding{51}       & \ding{55}           & \ding{55}                    \\
eSL-Net*           & ECCV 2020 & 1      & \ding{55} & \ding{51} & \ding{51}       & \ding{51}           & \ding{55}                    \\
EvUnroll           & CVPR 2022 & 1      & \ding{51} & \ding{51} & \ding{51}       & \ding{51}           & -                   \\
TimeLens           & CVPR 2021 & 2      & \ding{51} & \ding{51} & \ding{55}       & \ding{55}           & \ding{51}                    \\
E-CIR              & CVPR 2022 & 1      & \ding{55} & \ding{51} & \ding{51}       & \ding{55}           & \ding{51}                    \\
VideoINR        &    CVPR 2022 & 2    & \ding{51}   & \ding{55} & \ding{55}       & \ding{55}           & \ding{51}          \\
EvShutter          & CVPR 2023 & 1      & \ding{55} & \ding{55} & \ding{51}       & \ding{51}           & \ding{55} \\
DeblurSR           & AAAI 2024 & 1      & \ding{55} & \ding{51} & \ding{51}       & \ding{55}           & \ding{51}                    \\
NEIR               & Arxiv 2023& 1      & \ding{51} & \ding{51} & \ding{51}       & \ding{51}           & \ding{51}     \\
Ours               & - & 1      & \ding{51} & \ding{51} & \ding{51}       & \ding{51}           & \ding{51}                   \\
\bottomrule
\end{tabular}
}
}
\end{table}

\begin{figure*}[h]
\centering
\includegraphics[width=\linewidth]{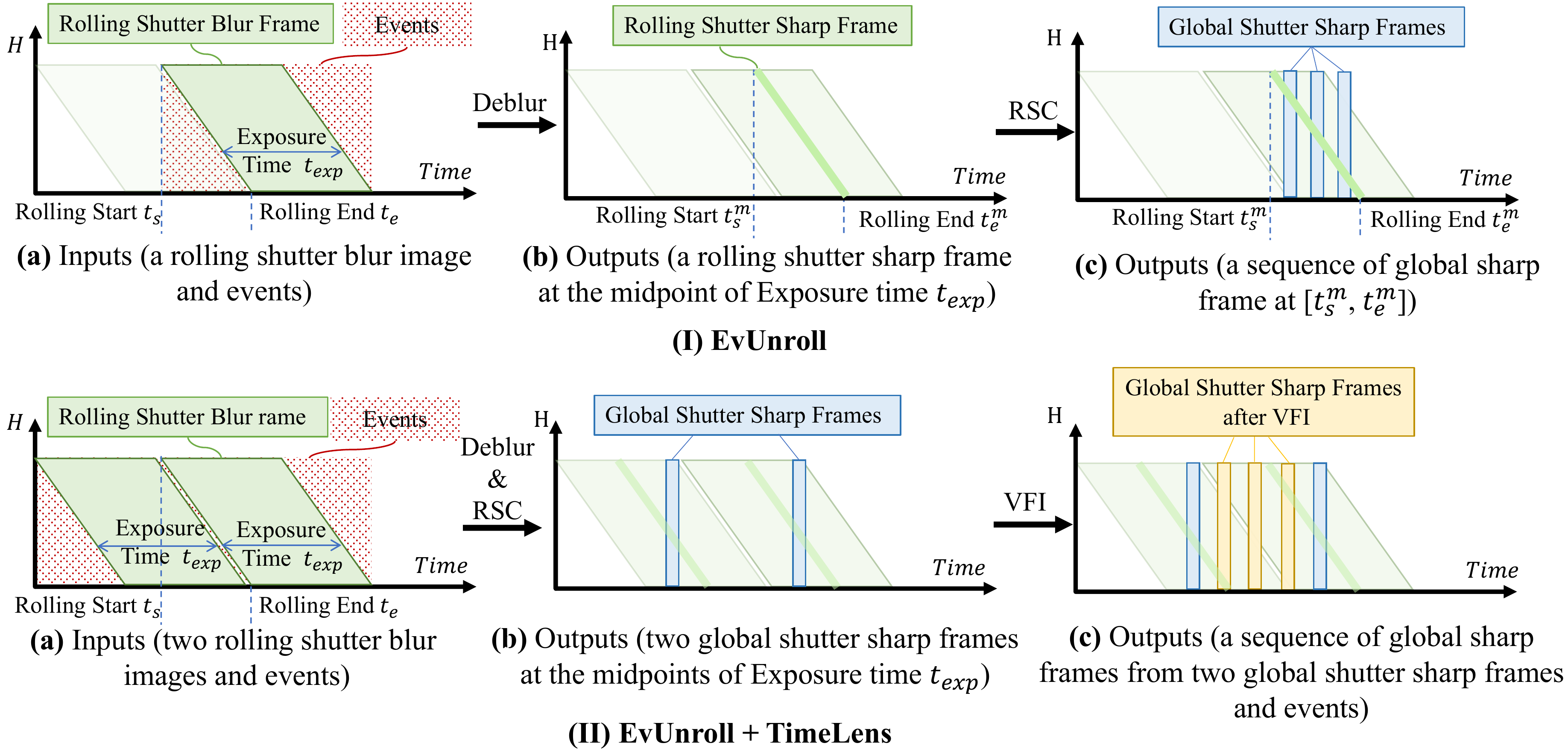}
\caption{Illustration of Experiment Settings of EvUnroll~\citep{zhou2022evunroll} and the combination of EvUnroll and TimeLens~\citep{tulyakov2021time}.}
\label{fig:experiment-setting}
\centering
\end{figure*}

In this section, we will explain the motivations for comparing EvUnroll~\citep{zhou2022evunroll} (event-guided RS correction) in the experiment that outputs a single GS sharp frame and comparing EvUnroll + Timelens~\citep{tulyakov2021time} (event-guided video frame interpolation) in the experiment that outputs a sequence of GS sharp frames.
Fig.~\ref{fig:experiment-setting} (I) shows the process of generating a sequence of global shutter sharp (GS) frames from a rolling shutter (RS) blur image and paired events by deblur and RS correction.
However, the deblur module in EvUnroll recovers the midpoint of the exposure time of each row~\citep{zhou2022evunroll}, as shown in Fig.~\ref{fig:experiment-setting} (b); furthermore, EvUnroll can only recover the GS sharp frames between the rolling start time $t_s^m$ and rolling end time $t_e^m$ of the reconstructed RS sharp frame, which can not output the arbitrary GS sharp frames during the whole exposure time of the RS blur frame.
Therefore, in the joint task of deblur and RS correction, EvUnroll can not realize arbitrary frame interpolation as shown in Tab.~\ref{tab:methods_comparison_table} and we combine EvUnroll and Timelens in the experiment outputting a sequence of GS sharp frames.
Specifically, we first generate two GS sharp frames with EvUnroll at the midpoint of the whole exposure time $t_{exp}$ from two RS blur frames and paired events, and then we use TimeLens to generate latent GS sharp frames with the input of two GS sharp frames and events, as shown in Fig.~\ref{fig:experiment-setting} (II).

\begin{figure*}[h]
\centering
\includegraphics[width=\linewidth]{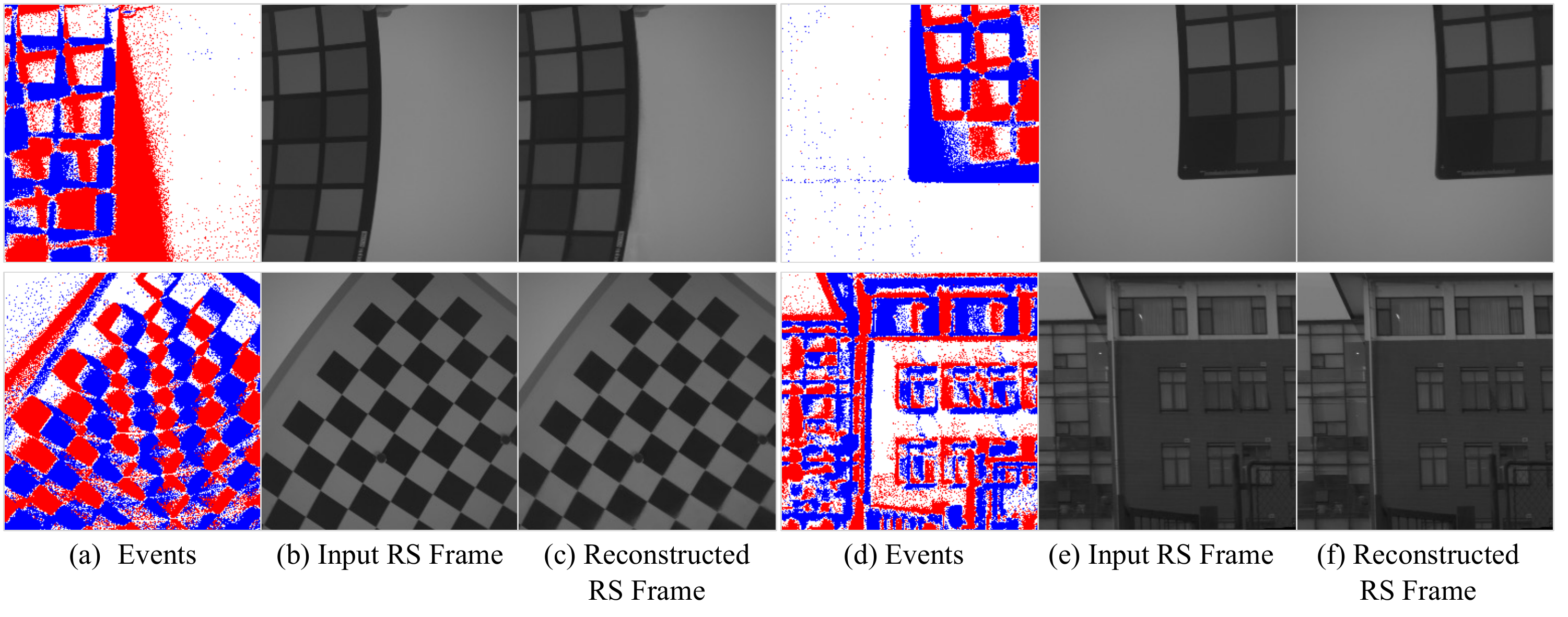}
\caption{Rolling shutter frame reconstruction visualization in real-world dataset.}
\label{fig:15-blur-reconstruction}
\centering
\end{figure*}

\begin{figure*}[h]
    \centering
    \includegraphics[width=1\linewidth]{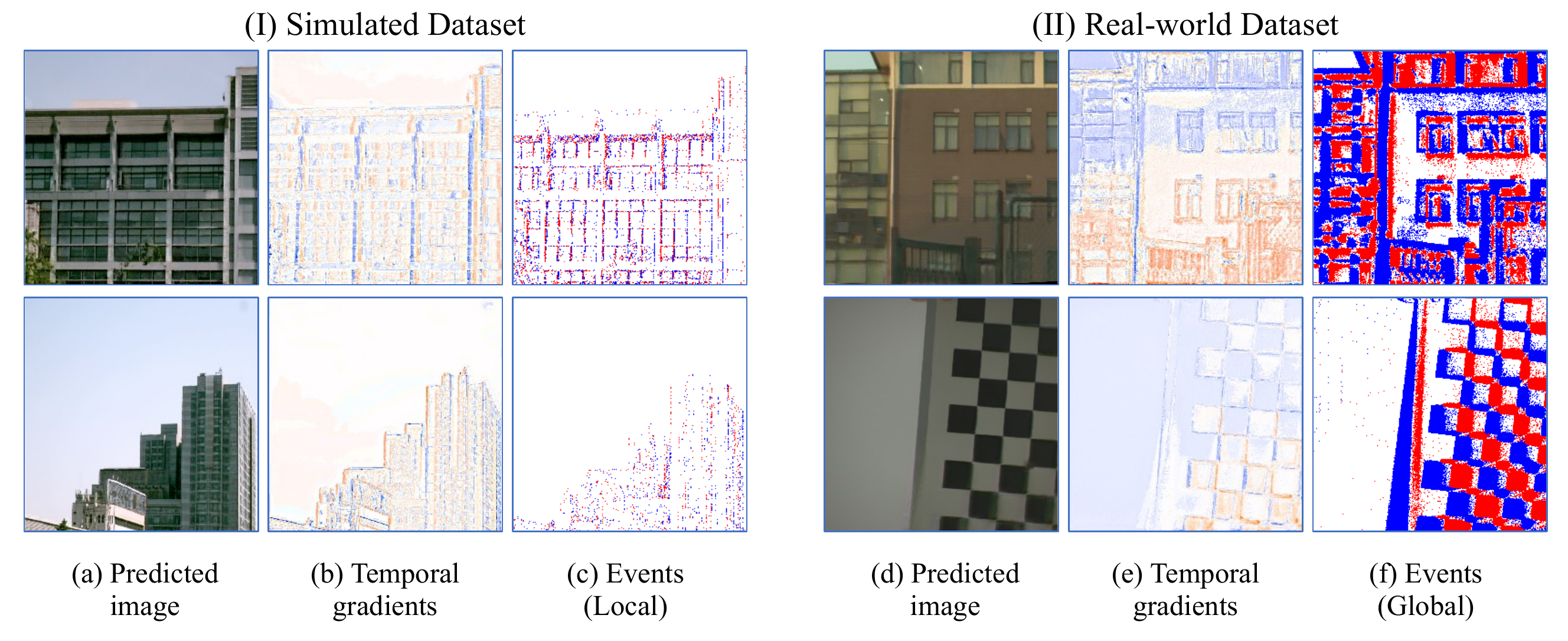}
    \caption{
    \small
    (I) and (I) show visualizations on simulated and real-world datasets, respectively.
    From left to right: the predicted images, temporal gradients ($\partial F(\bm{x},t,\theta) / \partial t$), and events.
    Orange and blue hues in the image signify positive and negative gradients, respectively.
    The color intensity is associated with the gradient value, with higher absolute values manifested by stronger colors.}
    \label{fig:temporal-gradients}
\end{figure*}

Compared with the latest research VideoINR \citep{chen2022videoinr}, our work differs in two aspects.
\textbf{a)} Different research questions:
While VideoINR tackles space-time super-resolution in the global shutter by introducing implicit neural representation (INR), our proposed method first simultaneously realizes RS correction, deblurring, and frame interpolation with INR.
\textbf{b)} Different methodologies:
\textit{i.} VideoINR consists of SpatialINR and TemporalINR, which are sequentially used to transfer the frame feature according to the spatial-temporal coordinate to achieve super-resolution and frame interpolation. However, SpatialINR, and TemporalINR cannot handle motion blur and rolling shutter distortion in the input frames.
\textit{ii.} In contrast, our approach develops a unified INR to simultaneously realize RS correction, deblurring, and frame interpolation. Especially, according to the principle of RS and GS images, we design Exposure Time Embedding enabling the generation of RS and GS images given the specific exposure time information, which is a feat unachievable by VideoINR due to its inconsideration towards RS distortion and blur.

\subsection{Visualization of Temporal Dimension Gradients:}

Fig.~\ref{fig:temporal-gradients} depicts the visualization of the gradients in the temporal dimension, demonstrating the successful training of the function $F(\bm{x},t,\theta)$.
Both the gradient visualization and events exhibit a similar intensity trend for $F(\bm{x},t,\theta)$ at the specified time $t$.
However, the gradient visualization appears smoother with more continuous edges.
This observation confirms that our method is capable of learning the high temporal resolution of intensity changes present in events, simultaneously filtering out noise.

\subsection{Exploring Why DeblurSR Appears to Correct Rolling Shutter:}
While the primary focus of the DeblurSR~\citep{song2023deblursr} study did not lie in the correction of the rolling shutter effect, we can observe a certain level of correction in the experiments, albeit accompanied by artifacts. We attribute this phenomenon to the fact that events themselves can be viewed as capturing a global shutter perspective. Consequently, the spiking representation learned by DeblurSR using events possesses the potential for rolling shutter correction.
The effectiveness of using events to learn implicit representation for rolling shutter correction is evident.

%
%

\end{document}